\documentclass{article} % For LaTeX2e
\usepackage{nips13submit_e, times, amsthm, amsmath, bbm, subfig, algorithm, algorithmic, amsfonts, amssymb, cite, graphicx, caption, enumerate, stmaryrd, ctable, hyperref, cancel, soul, footnote, xspace, placeins}

\newcommand{\x}{{\mathbf x}}
\newcommand{\bE}{{\mathbb E}}
\newcommand{\bR}{{\mathbb R}}

\newcommand{\cksny}{\texttt{XNV}\xspace}
\newcommand{\cksrff}{\texttt{XKS}\xspace}

\newcommand{\rffm}{\texttt{RFF}$_M$\xspace}
\newcommand{\rffM}{\texttt{RFF}$_{2M}$\xspace}

\newcommand{\sssl}{\texttt{SSSL}\xspace}
\newcommand{\ssslm}{\texttt{SSSL}$_M$\xspace}
\newcommand{\ssslM}{\texttt{SSSL}$_{2M}$\xspace}

\newcommand{\data}{{T}}
\newcommand{\ccH}{{\mathcal H}}

\newcommand{\wreg}{{\mathbf w}}
\newcommand{\wccaj}{{{\beta}}}
\newcommand{\wcca}{{\boldsymbol{\wccaj}}}

\DeclareMathOperator*{\argmin}{argmin}
\DeclareMathOperator*{\argmax}{argmax}

\newcommand{\loss}{\ell}

\newcommand{\savefootnote}[2]{\footnote{\label{#1}#2}}
\newcommand{\repeatfootnote}[1]{\textsuperscript{\ref{#1}}}

\newcommand{\tr}{^{\top}}

\newcommand{\cb}[1]{\left\{ {#1} \right\}}
\newcommand{\br}[1]{\left( {#1} \right)}
\newcommand{\sq}[1]{\left[ {#1} \right]}

\newcommand{\av}[1]{\left\langle{#1}\right\rangle}

\newcommand{\R}{{\mathbb R}} 
\newcommand{\N}{{\mathcal N}}

\newcommand{\y}{\mathbf{y}} 
\newcommand{\z}{\mathbf{z}} 
\newcommand{\K}{\mathbf{K}} 
 
\newcommand{\Id}{\mathbf {I}}
\newcommand{\bb}{{\mathbf{b}}} 
\newcommand{\wt}{{\mathbf{w}}}

\DeclareMathOperator{\lloss}{loss}

\newcommand{\Bt}{{\mathbf{B}}} 
 
\newcommand{\Dt}{{\mathbf{D}}}

\newcommand{\Kt}{{\mathbf{K}}}

\newcommand{\Vt}{{\mathbf{V}}}

\newcommand{\boldomega}{\boldsymbol{\omega}}

\newtheorem{thm}{Theorem}

\newtheorem{prop}[thm]{Proposition}

\newtheorem{defn}{Definition}
\newtheorem{assn}{Assumption}

\title{Correlated random features for \\fast semi-supervised learning}

\author{
Brian McWilliams \\
ETH Z\"urich, Switzerland\\
\texttt{brian.mcwilliams@inf.ethz.ch}
\And 
David Balduzzi \\
ETH Z\"urich, Switzerland\\
\texttt{david.balduzzi@inf.ethz.ch}
\And 
Joachim M. Buhmann\\
ETH Z\"urich, Switzerland\\
\texttt{jbuhmann@inf.ethz.ch}
}

\nipsfinalcopy % Uncomment for camera-ready version

\renewcommand{\algorithmicrequire}{\textbf{Input:}}
\renewcommand{\algorithmicensure}{\textbf{Output:}}

\begin{document}
\maketitle

\begin{abstract}
This paper presents Correlated Nystr\"om Views (\cksny), a fast semi-supervised algorithm for regression and classification. The algorithm draws on two main ideas. First, it generates two views consisting of computationally inexpensive random features. Second, multiview regression, using Canonical Correlation Analysis (CCA) on unlabeled data, biases the regression towards useful features. It has been shown that CCA regression can substantially reduce variance with a minimal increase in bias if the views contains accurate estimators. Recent theoretical and empirical work shows that regression with random features closely approximates kernel regression, implying that the accuracy requirement holds for random views. We show that \cksny consistently outperforms a state-of-the-art algorithm for semi-supervised learning: substantially improving predictive performance and reducing the variability of performance on a wide variety of real-world datasets, whilst also reducing runtime by orders of magnitude.
\end{abstract}

\section{Introduction}

As the volume of data collected in the social and natural sciences increases, the computational cost of learning from large datasets has become an important consideration. For learning non-linear relationships, kernel methods achieve excellent performance but na\"ively require operations cubic in the number of training points. 

Randomization has recently been considered as an alternative to optimization that, surprisingly, can yield comparable generalization performance at a fraction of the computational cost \cite{williams:01, rahimi:08}. Random features have been introduced to approximate kernel machines when the number of training examples is very large, rendering exact kernel computation intractable. Among several different approaches, the Nystr\"om method for low-rank kernel approximation \cite{williams:01} exhibits good theoretical properties and empirical performance \cite{yang:12,gittens:13,bach:13}.

A second problem arising with large datasets concerns obtaining \emph{labels}, which often requires a domain expert to manually assign a label to each instance which can be very expensive -- requiring significant investments of both time and money -- as the size of the dataset increases. Semi-supervised learning aims to improve prediction by extracting useful structure from the unlabeled data points and using this in conjunction with a function learned on a small number of labeled points. 

\paragraph{Contribution.}
This paper proposes a new semi-supervised algorithm for regression and classification, Correlated Nystr\"om Views (\cksny), that addresses both problems simultaneously. The method consists in essentially two steps. First, we construct two ``views'' using random features. We investigate two ways of doing so: one based on the Nystr\"om method and another based on random Fourier features (so-called kitchen sinks) \cite{rahimi:07, rahimi:08}. It turns out that the Nystr\"om method almost always outperforms Fourier features by a quite large margin, so we only report these results in the main text.

The second step, following \cite{kakade:07}, uses Canonical Correlation Analysis (CCA, \cite{hotelling:36, hardoon:04}) to bias the optimization procedure towards features that are correlated across the views. Intuitively, if both views contain accurate estimators, then penalizing uncorrelated features reduces variance without increasing the bias by much. Recent theoretical work by Bach \cite{bach:13} shows that Nystr\"om views can be expected to contain accurate estimators.

We perform an extensive evaluation of \cksny on 18 real-world datasets, comparing against a modified version of the \sssl (simple semi-supervised learning) algorithm introduced in \cite{ji:12}. We find that \cksny outperforms \sssl by around 10-15\% on average, depending on the number of labeled points available, see \S\ref{sec:results}. We also find that the performance of \cksny exhibits dramatically less variability than \sssl, with a typical reduction of 30\%.

We chose \sssl since it was shown in \cite{ji:12} to outperform a state of the art algorithm, Laplacian Regularized Least Squares \cite{belkin:06}. However, since \sssl does not scale up to large sets of unlabeled data, we modify \sssl by introducing a Nystr\"om approximation to improve runtime performance. This reduces runtime by a factor of $\times 1000$ on $N=10,000$ points, with further improvements as $N$ increases. Our approximate version of \sssl outperforms kernel ridge regression (KRR) by $>50\%$ on the 18 datasets on average, in line with the results reported in \cite{ji:12}, suggesting that we lose little by replacing the exact \sssl with our approximate implementation.

\paragraph{Related work.}
Multiple view learning was first introduced in the co-training method of \cite{blum:98} and has also recently been extended to unsupervised settings \cite{chaudhuri:09,mcwilliams:12}. Our algorithm builds on an elegant proposal for multi-view regression introduced in \cite{kakade:07}. Surprisingly, despite guaranteeing improved prediction performance under a relatively weak assumption on the views, CCA regression has not been widely used since its proposal -- to the best of our knowledge this is first empirical evaluation of multi-view regression's performance. A possible reason for this is the difficulty in obtaining naturally occurring data equipped with multiple views that can be shown to satisfy the multi-view assumption. We overcome this problem by constructing random views that satisfy the assumption by design.

\section{Method}
\label{sec:method}

%%%%%%%%%%%%%%%%%%%%%%%%%%%%%%%%%%%%%%%%%%%%%%%%%%%%%%%%%%%%%%%%%%%%%%%%%%%%%%%%
%%%%%%%%%%%%%%%%%%%%%%%%%%%%%%%%%%%%%%%%%%%%%%%%%%%%%%%%%%%%%%%%%%%%%%%%%%%%%%%%
\iffalse
\paragraph{Notation (for our use)}

$N$ - total number of labeled and unlabeled training points.

$n$ - number of labeled training points.

$D$ - number of features.

$M$ - number of random features.

$\Kt$ - kernel matrix.

$\kappa$ - kernel function.

$\lambda$ is either the regularizer or the CCA eigenvalue (we need to decide, maybe with subscripts?)

Lets not have the explicit dependence of $\z$ on $\x$. $\z(\x)$ gets real messy real fast.

$\wreg$ - regression coefficients

$\wcca$ - CCA regression coefficients
\fi
%%%%%%%%%%%%%%%%%%%%%%%%%%%%%%%%%%%%%%%%%%%%%%%%%%%%%%%%%%%%%%%%%%%%%%%%%%%%%%%%
%%%%%%%%%%%%%%%%%%%%%%%%%%%%%%%%%%%%%%%%%%%%%%%%%%%%%%%%%%%%%%%%%%%%%%%%%%%%%%%%

This section introduces \cksny, our semi-supervised learning method. The method builds on two main ideas. First, given two equally useful but sufficiently different views on a dataset, penalizing regression using the canonical norm (computed via CCA), can substantially improve performance \cite{kakade:07}. The second is the Nystr\"om method for constructing random features \cite{williams:01}, which we use to construct the views.

%In this section describe the multi-view CCA approach to semi-supervised learning as well as the idea of random feature construction for learning non-linear prediction functions. The multi-view setting assumes that two sets of features, or ``views" are collected on the same set of labels. A regression function can then be learned which weights features more heavily if they are strongly correlated between the views. Finally we present our algorithm which combines two randomly generated views for semi-supervised learning.  

%%%%%%%%%%%%%%%%%%%%%%%%%%%%%%%%%%%%%%%%%%%%%%%%%%%%%%%%%%%%%%%%%%%%%%%%%%%%%%%%
\subsection{Multi-view regression}

%\textcolor{red}{lets deal with random variables $x^{(1)}$ and $x^{(2)}$. We use the subscript $j$ for indexing features.}
Suppose we have data $\data=\big((\x_1,y_1),\ldots, (\x_n,y_n)\big)$ for $\x_i\in \bR^D$ and $y_i\in \bR$, sampled according to joint distribution $P(\x,y)$. Further suppose we have two \emph{views} on the data
\begin{equation*}
	\z^{(\nu)}: \bR^D \longrightarrow \ccH^{(\nu)}=\bR^M:\x\mapsto \z^{(\nu)}(\x)=:\z^{(\nu)}\quad\text{ for }\nu\in\{1,2\}.
\end{equation*}

We make the following assumption about linear regressors which can be learned on these views.

\begin{assn}[Multi-view assumption  \cite{kakade:07}]\label{as:mv}
	Define mean-squared error loss function $\loss(g,\x,y)=(g(\x)-y)^2$ and let $\lloss(g):=\bE_P \loss(g(\x),y)$. Further let $L(Z)$ denote the space of linear maps from a linear space $Z$ to the reals, and define:
	\begin{align*}
		f^{(\nu)} := \argmin_{g\in L(\ccH^{(\nu)})} \lloss(g)
		\; \text{ for }\nu\in\{1,2\}
		\quad{\text and }\quad
		f := \argmin_{g\in L(\ccH^{(1)}\oplus \ccH^{(2)})} \lloss(g).
	\end{align*}
	The multi-view assumption is that 
	\begin{equation}
		\label{a:mv}
		\lloss\left(f^{(\nu)}\right) - \lloss(f) \leq \epsilon
		\quad \text{ for }\nu\in\{1,2\}.
	\end{equation}
\end{assn}
In short, the best predictor in each view is within $\epsilon$ of the best overall predictor. %This implies that neither view contains substantially more useful information than the other. In fact, \cite{kakade:07} make a stronger assumption -- from which the above follows as a consequence. However, Eq.~\eqref{a:mv} suffices to derive their results on multi-view regression. 

\paragraph{Canonical correlation analysis.}
Canonical correlation analysis \cite{hotelling:36, hardoon:04} extends principal component analysis (PCA) from one to two sets of variables. CCA finds bases for the two sets of variables such that the correlation between projections onto the bases are maximized. 

The first pair of canonical basis vectors, $\br{\bb_1^{(1)}, \bb_1^{(2)}}$ is found by solving:
\begin{align}
	\argmax_{\bb^{(1)},\bb^{(2)} \in \R^{M} } \text{corr} \br{\bb^{(1)\top} \z^{(1)}, \bb^{(2)\top} \z^{(2)} } .
\end{align}

Subsequent pairs are found by maximizing correlations subject to being orthogonal to previously found pairs. The result of performing CCA is two sets of bases, $\Bt^{(\nu)}=\sq{\bb_1^{(\nu)},\ldots,\bb_M^{(\nu)}}$ for $\nu\in\{1,2\}$, such that the projection of $\z^{(\nu)}$ onto $\Bt^{(\nu)}$ which we denote $\bar{\z}^{(\nu)}$ satisfies
\begin{enumerate}
	\item \emph{Orthogonality:}
		$\bE_\data\big[\bar{\z}^{(\nu) \top}_j \bar{\z}^{(\nu)}_k]=\delta_{jk}$,			
	where $\delta_{jk}$ is the Kronecker delta, and
	\item \emph{Correlation:}
		$\bE_\data\big[\bar{\z}^{(1) \top}_j \bar{\z}^{(2)}_k\big] = \lambda_j\cdot \delta_{jk}$
	where w.l.o.g. we assume $1\geq \lambda_1\geq\lambda_2\geq\cdots\geq0$. 
\end{enumerate}
$\lambda_j$ is referred to as the $j^{th}$ \emph{canonical correlation coefficient}.

\begin{defn}[canonical norm]
Given vector $\bar{\z}^{(\nu)}$ in the canonical basis, define its \emph{canonical norm} as
\begin{equation*}
	\|\bar{\z}^{(\nu)}\|_{CCA} := \sqrt{\sum_{j=1}^D \frac{1-\lambda_j}{\lambda_j}\left(\bar{z}^{(\nu)}_j\right)^2} .
\end{equation*}
\end{defn}

\paragraph{Canonical ridge regression.} Assume we observe $n$ pairs of views coupled with real valued labels $\cb{\z_i^{(1)},\z_i^{(2)},y_i}_{i=1}^n$, canonical ridge regression
finds coefficients $\widehat{\wcca}^{(\nu)} = \sq{\widehat{\wccaj}_1^{(\nu)},\ldots,\widehat{\wccaj}_M^{(\nu)}}\tr $ such that
\begin{equation}
	\widehat{\wcca}^{(\nu)}:=\argmin_{\wcca}
	\frac{1}{n}\sum_{i=1}^n \left(y_i-{\wcca}^{(\nu) ~ \top} \bar{\z}^{(\nu)}_i\right)^2 + \|\wcca^{(\nu)}\|_{CCA}^2.
\end{equation}
The resulting estimator, referred to as the \emph{canonical shrinkage estimator}, is
\begin{equation}
	\label{e:shrinkage}
	\widehat{\wccaj}^{(\nu)}_j = \frac{\lambda_j}{n}\sum_{i=1}^n \bar{\z}^{(\nu)}_{i,j}y_i .
\end{equation}

Penalizing with the canonical norm biases the optimization towards features that are highly correlated across the views. Good regressors exist in both views by Assumption~\ref{a:mv}. Thus, intuitively, penalizing uncorrelated features significantly reduces variance, without increasing the bias by much. More formally:

\begin{thm}[canonical ridge regression, \cite{kakade:07}]\label{t:kakade}
	Assume $\bE[y^2|\x]\leq1$ and that Assumption~\ref{as:mv} holds. Let $f^{(\nu)}_{\widehat{\wcca}}$ denote the estimator constructed with the canonical shrinkage estimator, Eq.~\eqref{e:shrinkage}, on training set $\data$, and let $f$ denote the best linear predictor across both views. For $\nu\in\{1,2\}$ we have
	\begin{equation*}
		\bE_\data[\lloss(f_{\widehat{\wcca}}^{(\nu)})] - \lloss(f) \leq 5\epsilon + \frac{\sum_{j=1}^M \lambda_j^2}{n}
	\end{equation*}
	where the expectation is with respect to training sets $\data$ sampled from $P(\x,y)$.
\end{thm}

The first term, $5\epsilon$, bounds the bias of the canonical estimator, whereas the second, $\frac{1}{n}\sum \lambda_j^2$ bounds the variance. The $\sum \lambda_j^2$ can be thought of as a measure of the ``intrinsic dimensionality'' of the unlabeled data, which controls the rate of convergence. If the canonical correlation coefficients decay sufficiently rapidly, then the increase in bias is more than made up for by the decrease in variance.

%%%%%%%%%%%%%%%%%%%%%%%%%%%%%%%%%%%%%%%%%%%%%%%%%%%%%%%%%%%%%%%%%%%%%%%%%%%%%%%%
\subsection{Constructing random views}
\label{sec:rv} 
 
We construct two views satisfying Assumption~\ref{a:mv} in expectation, see Theorem~\ref{t:bach} below. To ensure our method scales to large sets of unlabeled data, we use random features generated using the Nystr\"om method \cite{williams:01}.

Suppose we have data $\{\x_i\}_{i=1}^N$. When $N$ is very large, constructing and manipulating the $N\times N$ Gram matrix  $\sq{\Kt}_{ii'} = \av{\phi(\x_i), \phi(\x_{i'})}=\kappa(\x_i,\x_{i'})$ is computationally expensive. Where here, $\phi(\x)$ defines a mapping from $\R^D$ to a high dimensional feature space and $\kappa(\cdot,\cdot)$ is a positive semi-definite kernel function.% The feature function $\phi(\x_i)$ defines a mapping from $\R^D$ to a higher dimensional space which is typically not feasible to compute explicitly. Similarly, evaluating the positive semi-definite kernel function $\kappa(\x_i,\x_{i'})$ on all pairs of points is expensive.

The idea behind random features is to instead define a lower-dimensional mapping, $\z(\x_i): \R^{D} \rightarrow \R^{M}$ through a random sampling scheme such that $\sq{\Kt}_{ii'}\approx\z(\x_i)\tr\z(\x_{i'})$ \cite{drineas:05,rahimi:07}. Thus, using random features, non-linear functions in $\x$ can be learned as linear functions in $\z(\x)$ leading to significant computational speed-ups. Here we give a brief overview of the Nystr\"om method, which uses random subsampling to approximate the Gram matrix.

\paragraph{The Nystr\"om method.} Fix an $M\ll N$ and randomly (uniformly) sample a subset $\mathcal{M}=\{\hat{\x}_i\}_{i=1}^M$ of $M$ points from the data $\{\x_i\}_{i=1}^N$. Let $\widehat{\K}$ denote the Gram matrix $[\widehat{\K}]_{ii'}$ where $i,i'\in {\mathcal M}$. The Nystr\"om method \cite{williams:01,yang:12} constructs a low-rank approximation to the Gram matrix as 
\begin{equation}
	\Kt \approx  \tilde{\K}:=\sum_{i=1}^N \sum_{i'=1}^N \sq{\kappa(\x_i,\hat{\x}_1),\ldots,\kappa(\x_i,\hat{\x}_M)}  	
	\widehat{\Kt}^{\dagger}\sq{\kappa(\x_{i'},\hat{\x}_1),\ldots,\kappa(\x_{i'},\hat{\x}_M)} \tr,
	\label{eq:nystromapprox}
\end{equation}
where $\widehat{\Kt}^{\dagger}\in\R^{M\times M}$ is the pseudo-inverse of $\widehat{\K}$. Vectors of random features can be constructed as 
$$
\z(\x_i) = \widehat{\Dt}^{-1/2}\widehat{\Vt}\tr  \sq{\kappa(\x_i,\hat{\x}_1),\ldots,\kappa(\x_i,\hat{\x}_M)}\tr,
$$ 
where the columns of $\widehat{\Vt}$ are the eigenvectors of $\widehat{\Kt}$ with $\widehat{\Dt}$ the diagonal matrix whose entries are the corresponding eigenvalues. Constructing features in this way reduces the time complexity of learning a non-linear prediction function from $O(N^3)$ to $O(N)$ \cite{drineas:05}.

An alternative perspective on the Nystr\"om approximation, that will be useful below, is as follows. Consider integral operators 
\begin{equation}
	\label{e:LN}
	L_N[f](\cdot) := \frac{1}{N}\sum_{i=1}^N \kappa(\x_i,\cdot)f(\x_i) \quad\text{ and }\quad
	L_M[f](\cdot) := \frac{1}{M}\sum_{i=1}^M \kappa(\x_i,\cdot)f(\x_i),
\end{equation}
and introduce Hilbert space $\hat{\ccH} = \text{span} \left\{\hat{\varphi}_1, \ldots, \hat{\varphi}_r\right\}$ where $r$ is the rank of $\widehat{\K}$ and the $\hat{\varphi}_i$ are the first $r$ eigenfunctions of $L_M$. Then the following proposition shows that using the Nystr\"om approximation is equivalent to performing linear regression in the feature space (``view'') $\z:{\mathcal X}\rightarrow \hat{\ccH}$ spanned by the eigenfunctions of linear operator $L_M$ in Eq.~\eqref{e:LN}:

\begin{prop}[random Nystr\"om view, \cite{yang:12}]\label{t:yang}
	Solving 
	\begin{equation}
		\label{e:fspace1}
		\min_{\wt\in\bR^r} \frac{1}{N}\sum_{i=1}^N \loss(\wt \tr \z(\x_i),y_i) + \frac{\gamma}{2}\|\wt\|^2_2
	\end{equation}
	is equivalent to solving
	\begin{equation}
		\label{e:fspace2}
		\min_{f\in \hat{\ccH}} \frac{1}{N}\sum_{i=1}^N \loss(f(\x_i),y_i) + \frac{\gamma}{2} \|f\|^2_{\ccH_\kappa}.
	\end{equation}
\end{prop}

%%%%%%%%%%%%%%%%%%%%%%%%%%%%%%%%%%%%%%%%%%%%%%%%%%%%%%%%%%%%%%%%%%%%%%%%%%%%%%%%
\subsection{The proposed algorithm: Correlated Nystr\"om Views (\cksny)}

Algorithm \ref{alg:xnv} details our approach to semi-supervised learning based on generating two views consisting of Nystr\"om random features and penalizing features which are weakly correlated across views. The setting is that we have labeled data $\{\x_i,y_i \}_{i=1}^n$ and a large amount of unlabeled data $\{\x_i\}_{i=n+1}^N$.

\begin{algorithm}[htp]
\caption{\texttt{Correlated Nystr\"om Views (\cksny)}.\label{alg:xnv}}
	\algorithmicrequire\; Labeled data: $\{\x_i,y_i\}_{i=1}^n$ and unlabeled data: $\{\x_i\}_{i=n+1}^{N}$
  \begin{algorithmic}[1]
    \STATE {\bf \emph{Generate features.}} Sample $\hat{\x}_1,\ldots,\hat{\x}_{2M}$ uniformly from the dataset, compute the eigendecompositions of the sub-sampled kernel matrices $\hat{\Kt}^{(1)}$ and $\hat{\Kt}^{(2)}$ which are constructed from the samples $1,\ldots,M$ and $M+1,\ldots,2M$ respectively, and featurize the input: 
        \begin{equation*}
    		\quad \z^{(\nu)}(\x_i)\leftarrow \hat{\Dt}^{(\nu), -1/2} \hat{\Vt}^{(\nu)\top} \sq{\kappa(\x_i,\hat{\x}_1),\ldots, \kappa(\x_i,\hat{\x}_M) }\tr
			\text{ for }\nu\in\{1,2\}. 
        \end{equation*}

	\STATE {\bf \emph{Unlabeled data.}} Compute CCA bases $\Bt^{(1)}$, $\Bt^{(2)}$ and canonical correlations $\lambda_1,\ldots,\lambda_M$ for the two views and set
    $\bar{\z}_i \leftarrow \Bt^{(1)}\z^{(1)}(\x_i).$
    \STATE {\bf \emph{Labeled data.}} Solve
    \begin{equation}
%<<<<<<< .mine
%   \widehat{\wcca} =  \arg\min_{\wcca} \frac{1}{n} \sum_{i=1}^n \loss \br{\wcca \tr \bar{\z}_i, y_i} + \Vert \wcca \Vert^2_{CCA} + \gamma \Vert \wcca \Vert_2^2 ~.
%=======
   \widehat{\wcca} =  \argmin_{\wcca} \frac{1}{n} \sum_{i=1}^n \loss \br{\wcca \tr \bar{\z}_i, y_i} + \Vert \wcca \Vert^2_{CCA} + \gamma \Vert \wcca \Vert_2^2 ~.
%>>>>>>> .r20022
    \label{eq:minCCA}
    \end{equation}    
  \end{algorithmic}
  \algorithmicensure\; $\widehat{\wcca}$
\end{algorithm}

Step 1 generates a set of random features. The next two steps implement multi-view regression using the randomly generated views $\z^{(1)}(\x)$ and $\z^{(2)}(\x)$. Eq.~\eqref{eq:minCCA} yields a solution for which unimportant features are heavily downweighted in the CCA basis \emph{without} introducing an additional tuning parameter. The further penalty on the $\ell_2$ norm (in the CCA basis) is introduced as a practical measure to control the variance of the estimator $\widehat{\wcca}$ which can become large if there are many highly correlated features (i.e. the ratio $\frac{1-\lambda_j}{\lambda_j} \approx 0$ for large $j$). In practice most of the shrinkage is due to the CCA norm: cross-validation obtains optimal values of $\gamma$ in the range $[0.00001,0.1]$.

\paragraph{Computational complexity.}
\cksny is extremely fast. Nystr\"om sampling, step 1, reduces the $O(N^3)$ operations required for kernel learning to $O(N)$. Computing the CCA basis, step 2, using standard algorithms is in $O(NM^2)$. However, we reduce the runtime to $O(NM)$ by applying a recently proposed randomized CCA algorithm of \cite{avron:13}. Finally, step 3 is a computationally cheap linear program on $n$ samples and $M$ features.  

\paragraph{Performance guarantees.}
The quality of the kernel approximation in \eqref{eq:nystromapprox} has been the subject of detailed study in recent years leading to a number of strong empirical and theoretical results \cite{yang:12,gittens:13,bach:13,drineas:05}. Recent work of Bach \cite{bach:13} provides theoretical guarantees on the quality of Nystr\"om estimates in the fixed design setting that are relevant to our approach.\footnote{Extending to a random design requires techniques from \cite{hsu:12}.}

\begin{thm}[Nystr\"om generalization bound, \cite{bach:13}]
	\label{t:bach}
	Let $\xi\in \bR^N$ be a random vector with finite variance and zero mean, $\y=\sq{y_1,\ldots,y_N}\tr$, and define smoothed estimate $\hat{\y}_{\text{kernel}}:= (\K+N\gamma\Id)^{-1}\K(\y+\xi)$ and smoothed Nystr\"om estimate $\hat{\y}_{\text{Nystr\"om}}:=(\tilde{\K}+N\gamma \Id)^{-1}\tilde{\K}(\y+\xi)$, both computed by minimizing the MSE with ridge penalty $\gamma$. Let $\eta\in(0,1)$. For sufficiently large $M$ (depending on $\eta$, see \cite{bach:13}), we have
	\begin{equation*}
		\bE_{\mathcal M}\bE_\xi\left[\|\y-\hat{\y}_{\text{Nystr\"om}}\|^2_2\right]
		\leq (1+4\eta) \cdot \bE_\xi\left[\|\y-\hat{\y}_{\text{kernel}}\|^2_2\right]
	\end{equation*}
	where $\bE_{\mathcal M}$ refers to the expectation over subsampled columns used to construct $\tilde{\K}$.
\end{thm}

In short, the best smoothed estimators in the Nystr\"om views are close to the optimal smoothed estimator. Since the kernel estimate is consistent, $\lloss(f)\rightarrow0$ as $n\rightarrow\infty$. Thus, Assumption~\ref{a:mv} holds in expectation and the generalization performance of \cksny is controlled by Theorem~\ref{t:kakade}.

\paragraph{Random Fourier Features.}
An alternative approach to constructing random views is to use Fourier features instead of Nystr\"om features in Step 1. We refer to this approach as {Correlated Kitchen Sinks (\cksrff)} after \cite{rahimi:08}. It turns out that the performance of \cksrff is consistently worse than \cksny, in line with the detailed comparison presented in \cite{yang:12}. We therefore do not discuss Fourier features in the main text, see \S\ref{s:rff} for details on implementation and experimental results.

%%%%%%%%%%%%%%%%%%%%%%%%%%%%%%%%%%%%%%%%%%%%%%%%%%%%%%%%%%%%%%%%%%%%%%%%%%%%%%%%
\subsection{A fast approximation to \sssl}
\label{s:sssl}

The \sssl (simple semi-supervised learning) algorithm proposed in \cite{ji:12} finds the first $s$ eigenfunctions $\phi_i$ of the integral operator $L_N$ in Eq.~\eqref{e:LN} and then solves 
\begin{equation}
	\label{e:sssl}
	\argmin_{\wt\in \bR^s} \sum_{i=1}^n\left(\sum_{j=1}^s w_j\phi_k(\x_i)-y_i\right)^2 , 
\end{equation}
where $s$ is set by the user. \sssl outperforms Laplacian Regularized Least Squares \cite{belkin:06}, a state of the art semi-supervised learning method, see \cite{ji:12}. It also has good generalization guarantees under reasonable assumptions on the distribution of eigenvalues of $L_N$. However, since \sssl requires computing the full $N\times N$ Gram matrix, it is extremely computationally intensive for large $N$. Moreover, tuning $s$ is difficult since it is discrete. 

We therefore propose \ssslm, an approximation to $\sssl$. First, instead of constructing the full Gram matrix, we construct a Nystr\"om approximation by sampling $M$ points from the labeled and unlabeled training set. Second, instead of thresholding eigenfunctions, we use the easier to tune ridge penalty which penalizes directions proportional to the inverse square of their eigenvalues \cite{dhillon:11}. 

As justification, note that Proposition~\ref{t:yang} states that the Nystr\"om approximation to kernel regression actually solves a ridge regression problem in the span of the eigenfunctions of $\hat{L}_M$. As $M$ increases, the span of $\hat{L}_M$ tends towards that of $L_N$ \cite{drineas:05}. We will also refer to the Nystr\"om approximation to \sssl using $2M$ features as \ssslM. See experiments below for further discussion of the quality of the approximation.

\section{Experiments}
\label{sec:results}

\paragraph{Setup.} We evaluate the performance of \cksny on 18 real-world datasets, see Table~\ref{tab:datasets}. The datasets cover a variety of regression (denoted by R) and two-class classification (C) problems. The \texttt{sarcos} dataset involves predicting the joint position of a robot arm; following convention we report results on the 1st, 5th and 7th joint positions. 

%\texttt{HIVa}, \texttt{ibn Sina}, \texttt{orange} and \texttt{sylva} are taken from the active learning challenge 

\begin{savenotes}
\begin{table}
\begin{center}
\small
\vspace{-5pt}
\setlength{\tabcolsep}{3.5pt}
\caption{Datasets used for evaluation.\label{tab:datasets}}
\vspace{-5pt}
\begin{tabular}{r l l r r | r l l r r}
\hline
Set & Name & Task &     N & D & Set & Name & Task &     N & D \\
\hline
1 &\texttt{abalone}\savefootnote{ft:uci}{{Taken from the UCI repository \url{http://archive.ics.uci.edu/ml/datasets.html}} }      &C &      $2,089$ & $6$  &   
10 &\texttt{elevators}\repeatfootnote{ft:lt}     &R &    $8,752$   & $18$    \\

2 &\texttt{adult}\repeatfootnote{ft:uci}       &C &     $32,561$ & $14$   & 
11 &\texttt{HIVa}\savefootnote{ft:al}{{Taken from %the active learning challenge
 \url{http://www.causality.inf.ethz.ch/activelearning.php}} }        &C &    $21,339$  & $1,617$   \\

3 &\texttt{ailerons}\savefootnote{ft:lt}{{Taken from \url{http://www.dcc.fc.up.pt/~ltorgo/Regression/DataSets.html}}}     &R &    $7,154$   & $40$    &
12 &\texttt{house}\repeatfootnote{ft:lt}       &R &    $11,392$  & $16$    \\

4 &\texttt{bank8}\repeatfootnote{ft:lt}        &C &    $4,096$   & $8$    &
13 &\texttt{ibn Sina}\repeatfootnote{ft:al}     &C &    $10,361$  & $92$    \\

5 &\texttt{bank32}\repeatfootnote{ft:lt}        &C &    $4,096$   & $32$    &
14 &\texttt{orange}\repeatfootnote{ft:al}      &C &    $25,000$  & $230$    \\

6 &\texttt{cal housing}\repeatfootnote{ft:lt}           &R &    $10,320$  & $8$  &  
{15} &{\texttt{sarcos 1}}\savefootnote{ft:gp}{Taken from \url{http://www.gaussianprocess.org/gpml/data/}}          &R &    $44,484$    & $21$    \\

7 &\texttt{census}\repeatfootnote{ft:uci}       &R &    $18,186$  & $119$   &
{16} &{\texttt{sarcos 5}}\repeatfootnote{ft:gp}         &R &    $44,484$    & $21$    \\

8 &\texttt{CPU}\repeatfootnote{ft:uci}          &R &    $6,554$   & $21$    &
17 &\texttt{sarcos 7}\repeatfootnote{ft:gp}    &R &    $44,484$  & $21$    \\

9 &\texttt{CT}\repeatfootnote{ft:uci}           &R &    $30,000$  & $385$   &
18 &\texttt{sylva}\repeatfootnote{ft:al}        &C &    $72,626$  & $216$    \\

\hline 
\end{tabular}
\vspace{-20pt}
\end{center}
\end{table}
\end{savenotes}

The \sssl algorithm was shown to exhibit state-of-the-art performance over fully and semi-supervised methods in scenarios where few labeled training examples are available \cite{ji:12}. 
However, as discussed in \S\ref{sec:rv}, due to its computational cost we compare the performance of \cksny to the Nystr\"om approximations \ssslm and \ssslM. %We refer to the approximation of \sssl with $M$ and $2M$ random features as \ssslm and \ssslM respectively. % \sssl was shown to exhibit state-of-the-art performance over fully and semi-supervised methods in scenarios where few labeled training examples are available \cite{ji:12}. 

We used a Gaussian kernel for all datasets. We set the kernel width, $\sigma$ and the $\ell_2$ regularisation strength, $\gamma$, for each method using 5-fold cross validation with $1000$ labeled training examples. We trained all methods using a squared error loss function, $\loss(f(\x_i),y_i) = (f(\x_i) - y_i)^2$, with $M=200$ random features, and $n=100,150,200,\ldots,1000$ randomly selected training examples.% Prediction errors were computed on an unseen test set. The mean and standard deviation of the prediction error over 100 repetitions of the experiments are reported. 

\paragraph{Runtime performance.} 
The \sssl algorithm of \cite{ji:12} is not computationally feasible on large datasets, since it has time complexity $O(N^3)$. For illustrative purposes, we report run times\footnote{Computed in Matlab 7.14 on a Core i5 with 4GB memory.} in seconds of the \sssl algorithm against \ssslm and \cksny on three datasets of different sizes.
%\begin{table}
\begin{center}
\vspace{-5pt}
\small
\setlength{\tabcolsep}{3.5pt}
%\caption{Datasets used for evaluation.\label{tab:datasets}}
\begin{tabular}{l r r r}
\hline
runtimes & \texttt{bank8} & \texttt{cal housing} & \texttt{sylva}
  \\
\hline
\sssl  & $72$s  & $2300$s & - \\
\ssslM & $0.3$s & $0.6$s  & $24$s \\  
\cksny & $0.9$s & $1.3$s  & $26$s \\
\hline 
\end{tabular}
\vspace{-5pt}
\end{center}
%\end{table}
For the \texttt{cal housing} dataset, \cksny exhibits an almost $1800\times$ speed up over \sssl. For the largest dataset, \texttt{sylva}, exact \sssl is computationally intractable. Importantly, the computational overhead of \cksny over \ssslM is small.

\paragraph{Generalization performance.}
We report on the prediction performance averaged over 100 experiments. For regression tasks we report on the mean squared error (MSE) on the testing set normalized by the variance of the test output. For classification tasks we report the percentage of the test set that was misclassified. 

The table below shows the improvement in performance of \cksny over \ssslm and \ssslM (taking whichever performs better out of $M$ or $2M$ on each dataset), averaged over all 18 datasets. Observe that \cksny is considerably more accurate and more robust than \ssslm. 

%%%% Table average %%%%%%%%%%%%%%%%%%%%%%%%%%%%%%%%%%%%%%%%%%%%%%%%%%
\begin{center}
{	\small
\begin{tabular}{l r r r r r}
\hline
\cksny vs \texttt{SSSL}$_{M/2M}$ & $n=100$ & $n=200$ & $n=300$ & $n=400$ & $n=500$ \\
\hline
Avg reduction in error  &  11\%  & 16\%  &  15\% &  12\% & 9\%  \\ 
Avg reduction in std err   &  15\%  & 30\%  &  31\% &  33\% & 30\% \\
\hline
\end{tabular}
}
\end{center}
The reduced variability is to be expected from Theorem~\ref{t:kakade}. 

\begin{figure}[htp]
\vspace{-20pt}
\begin{centering}
\subfloat[\texttt{adult}]{\includegraphics[width=0.33\columnwidth]{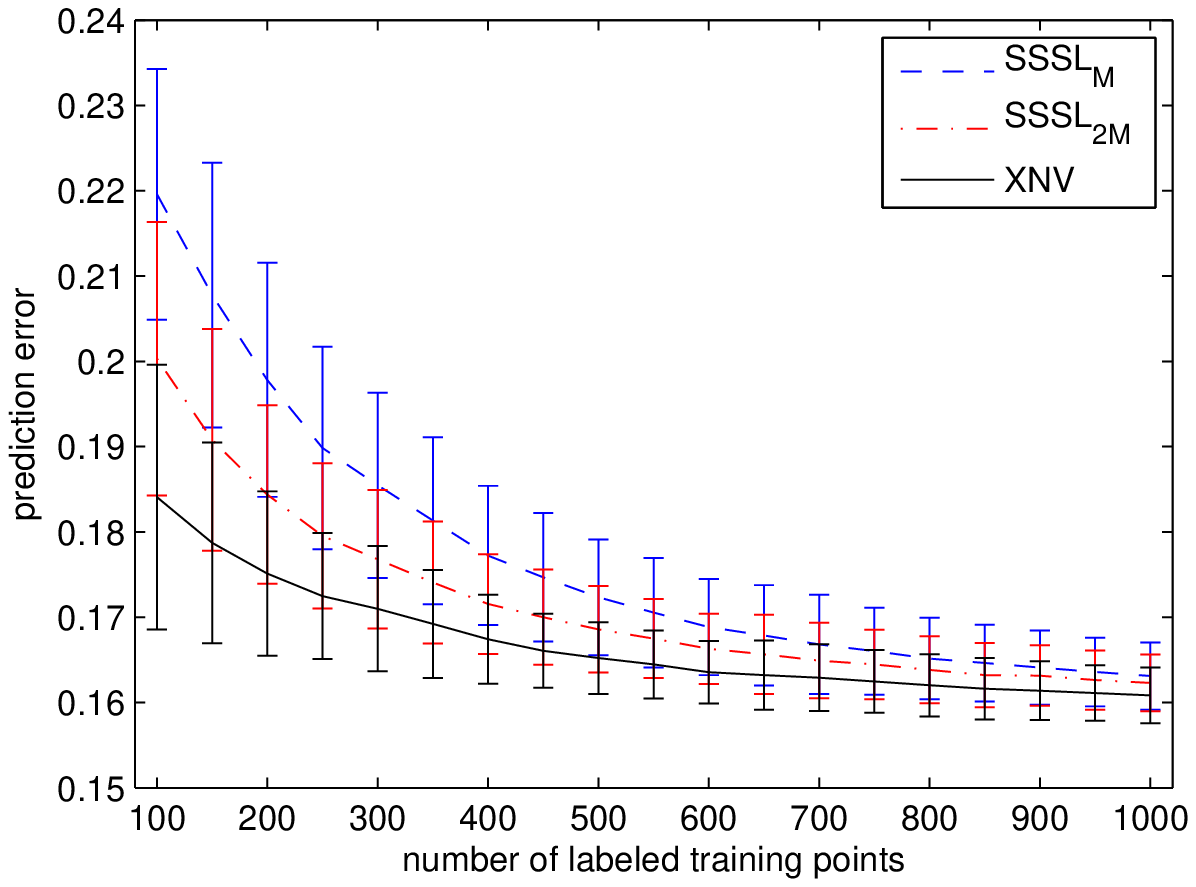} \label{fig:adult}} 
\subfloat[\texttt{cal housing}]{\includegraphics[width=0.33\columnwidth]{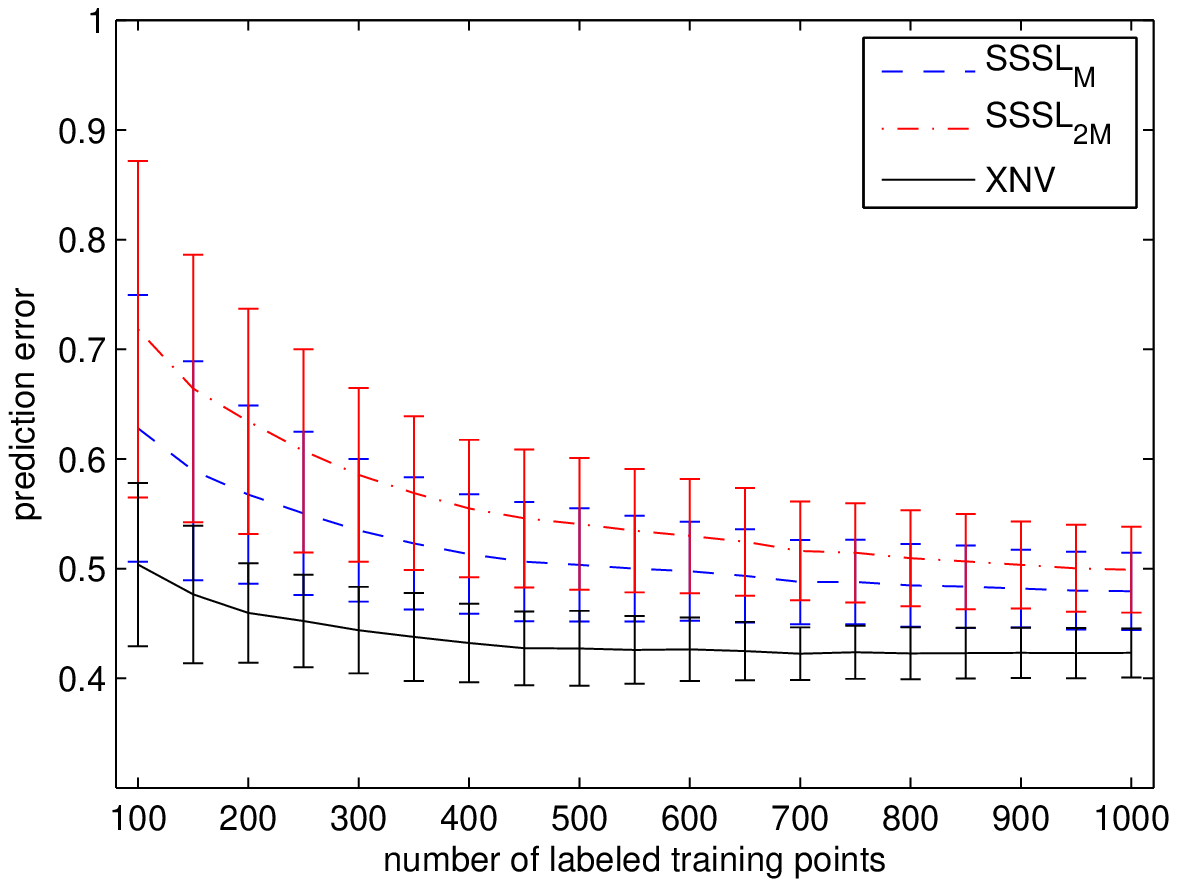} \label{fig:cal}}
\subfloat[\texttt{census}]{\includegraphics[width=0.33\columnwidth]{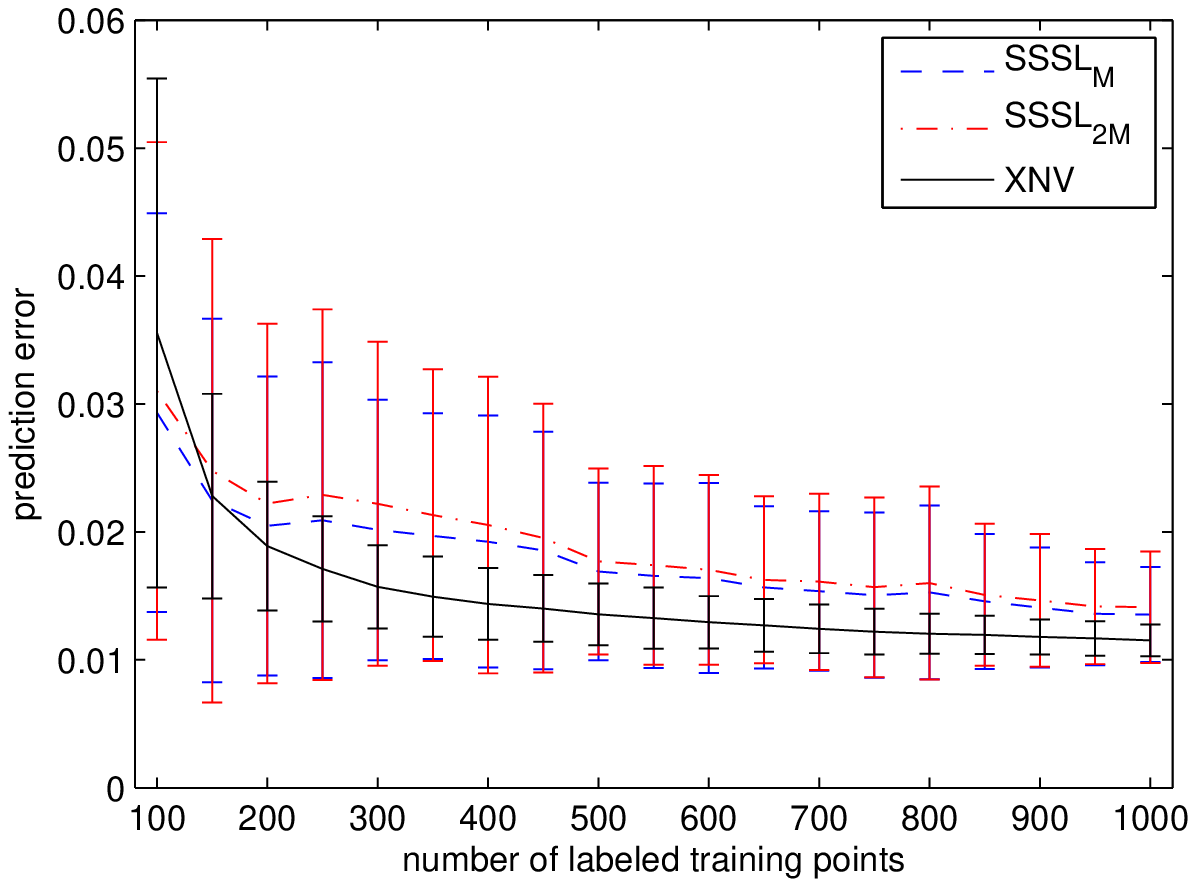} \label{fig:census}} 
\vspace{-10pt}
\subfloat[\texttt{elevators}]{\includegraphics[width=0.33\columnwidth]{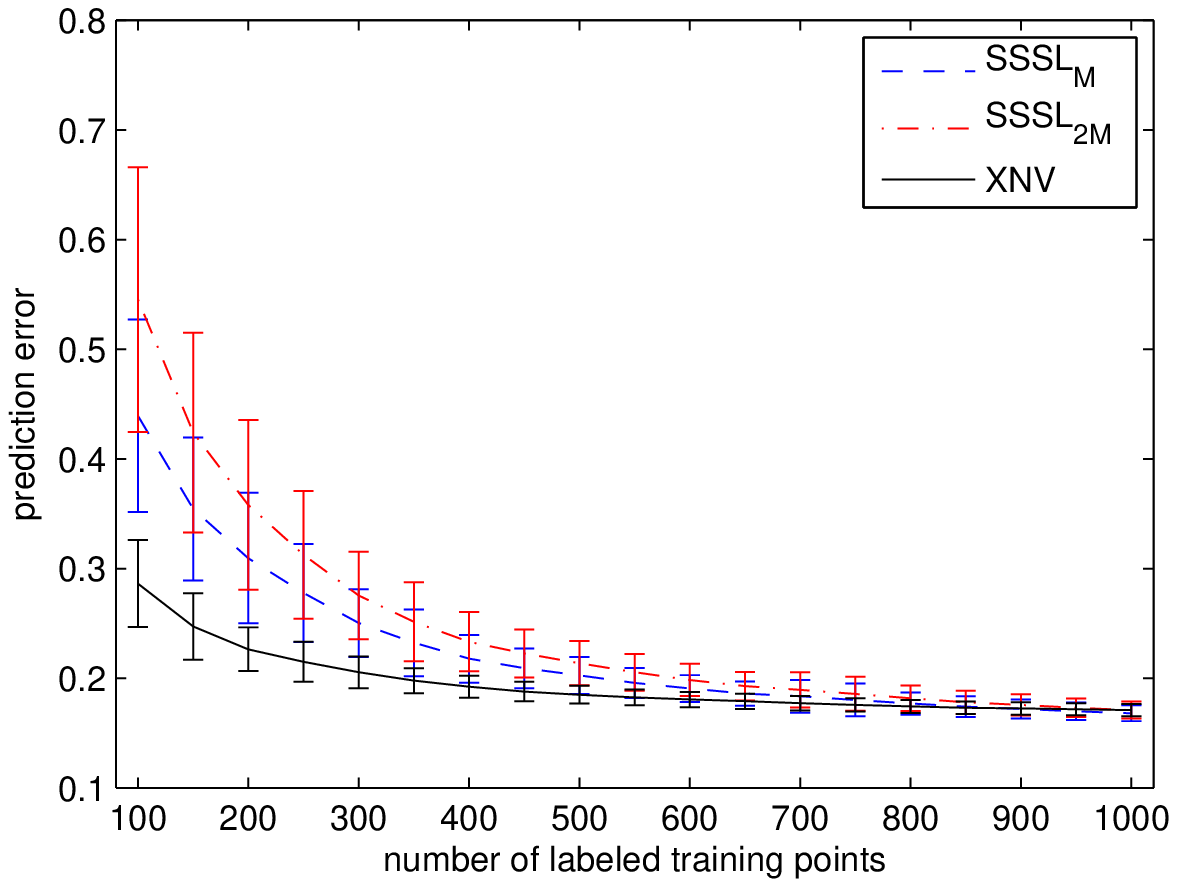} \label{fig:elevators}} 
\subfloat[\texttt{ibn Sina}]{\includegraphics[width=0.33\columnwidth]{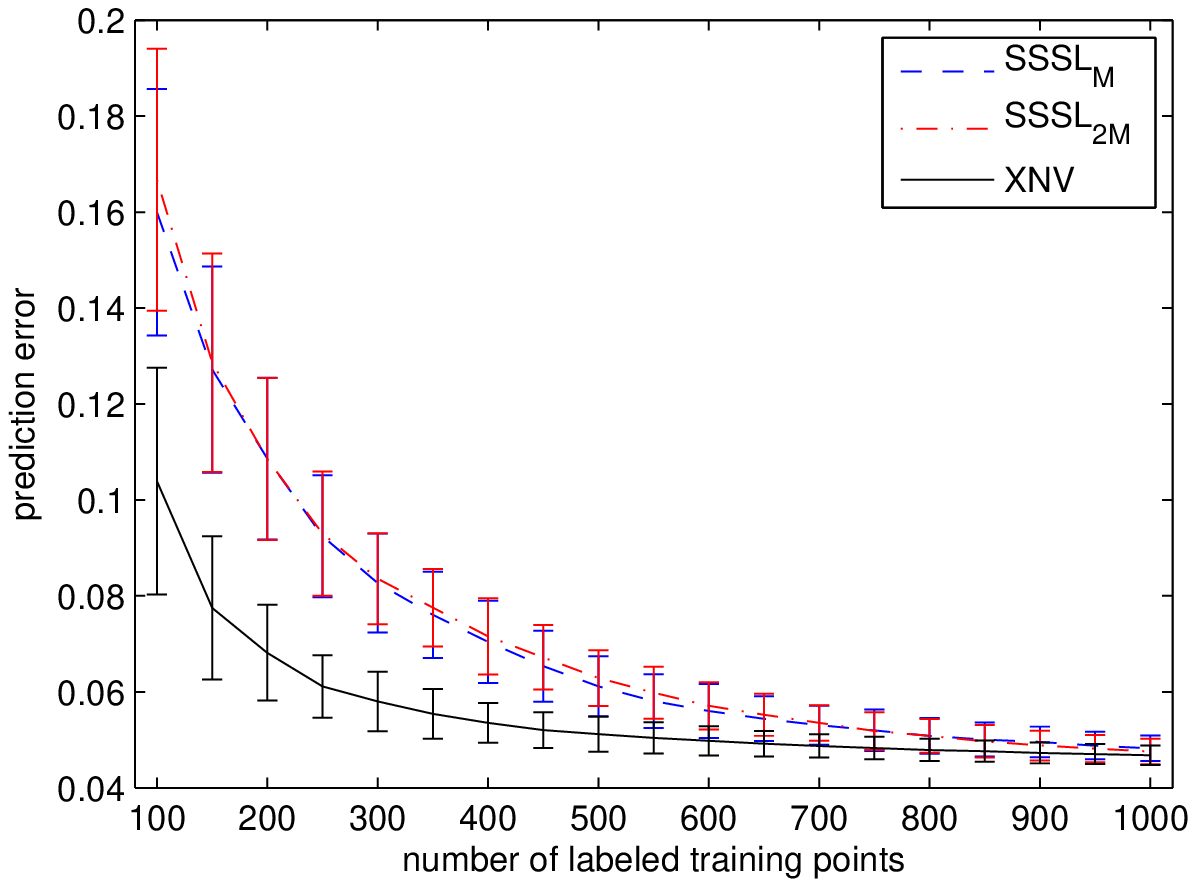} \label{fig:ibn_sina}} 
\subfloat[\texttt{sarcos 5}]{\includegraphics[width=0.33\columnwidth]{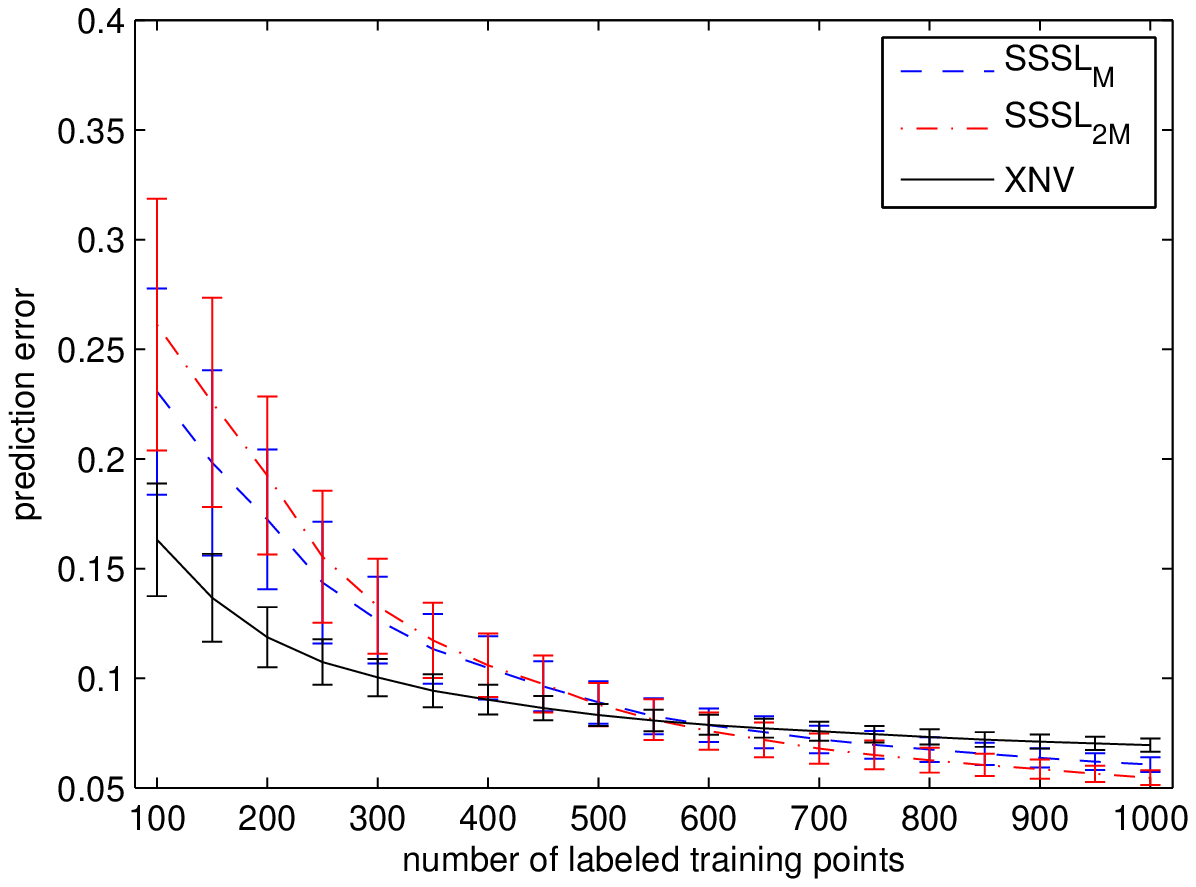} \label{fig:sarcos5}}
\vspace{-5pt}
\caption{Comparison of mean prediction error and standard deviation on a selection of datasets. \label{fig:res1}} 
\end{centering}
\end{figure}

Table~\ref{tab:results} presents more detailed comparison of performance for individual datasets when $n=200,400$. %We call the approximation to \sssl using $M$ and $2M$ samples as \ssslm and \ssslM respectively. 
The plots in Figure \ref{fig:res1} shows a representative comparison of mean prediction errors for several datasets when $n=100,\ldots,1000$. Error bars represent one standard deviation. Observe that \cksny almost always improves prediction accuracy and reduces variance compared with \ssslm and \ssslM when the labeled training set contains between 100 and 500 labeled points. A complete set of results is provided in \S\ref{s:xnvres}.

\paragraph{Discussion of \ssslm.}
Our experiments show that going from $M$ to $2M$ does not improve generalization performance in practice.  This suggests that when there are few labeled points, obtaining a more accurate estimate of the eigenfunctions of the kernel does not necessarily improve predictive performance. Indeed, when more random features are added, stronger regularization is required to reduce the influence of uninformative features, this also has the effect of downweighting informative features. This suggests that the low rank approximation \ssslm to $\sssl$ suffices.  

Finally, \S\ref{sec:krr} compares the performance of \ssslm and \cksny to fully supervised kernel ridge regression (KRR). We observe dramatic improvements, between 48\% and 63\%, consistent with the results observed in \cite{ji:12} for the exact \sssl algorithm.

\paragraph{Random Fourier features.}
Nystr\"om features significantly outperform Fourier features, in line with observations in \cite{yang:12}. The table below shows the relative improvement of \cksny over \cksrff:
%%%% Table average difference %%%%%%%%%%%%%%%%%%%%%%%%%%%%%%%%%%%%%%%%%%%%%%%%%%
\begin{center}
{\small
\begin{tabular}{l r r r r r}
\hline
\cksny vs \cksrff  & $n=100$ & $n=200$ & $n=300$ & $n=400$ & $n=500$ \\
\hline
Avg reduction in error  &  30\%  & 28\%  &  26\% &  25\% & 24\%  \\ 
Avg reduction in std err   &  36\%  & 44\%  &  34\% &  37\% & 36\% \\
\hline
\end{tabular}
}
\end{center}
Further results and discussion for \cksrff are included in the supplementary material.

%%%% Table average %%%%%%%%%%%%%%%%%%%%%%%%%%%%%%%%%%%%%%%%%%%%%%%%%%
%\begin{table}[htb]
%\begin{center}
%\vspace{-5pt}
%\setlength{\tabcolsep}{3.5pt}
%\small
%\caption{Average performance of \cksny across 18 datasets.\label{tab:average}}
%\vspace{-5pt}
%\begin{tabular}{l r r r r r}
%\hline
%& $n=100$ & $n=200$ & $n=300$ & $n=400$ & $n=500$ \\
%\hline
%Reduction in error  &  11\%  & 16\%  &  15\% &  12\% & 9\%  \\ 
%Reduction in s.d.   &  15\%  & 30\%  &  31\% &  33\% & 30\% \\
%\hline
%\end{tabular}
%\vspace{-15pt}
%\end{center}
%\end{table}

%%%% Table n=100 %%%%%%%%%%%%%%%%%%%%%%%%%%%%%%%%%%%%%%%%%%%%%%%%%%
\begin{table}[htp]
\begin{center}
\setlength{\tabcolsep}{3.0pt}
\small
\caption{Performance (normalized MSE/classification error rate). Standard errors in parentheses. \label{tab:results}}
\vspace{-5pt}
\begin{tabular}{l r r r | l r r r }
\hline
\multicolumn{1}{c}{set}&
\multicolumn{1}{c}{\ssslm}&
\multicolumn{1}{c}{\ssslM}&
\multicolumn{1}{c}{\cksny}
 & set &
\multicolumn{1}{c}{\ssslm}&
\multicolumn{1}{c}{\ssslM}&
\multicolumn{1}{c}{\cksny} \\ 
 %& \ssslm & \ssslM & \cksny \\
\hline
\multicolumn{8}{l}{$n=200$}\\
\hline
1& $0.054$ $(0.005)$& $0.055$ $(0.006)$& ${\bf 0.053}$ $({\bf 0.004})$  & 10& $0.309$ $(0.059)$& $0.358$ $(0.077)$& ${\bf 0.226}$ $({\bf 0.020})$  \\ 
2& $0.198$ $(0.014)$& $0.184$ $(0.010)$& ${\bf 0.175}$ $({\bf 0.010})$  & 11& $0.146$ $(0.048)$& $0.072$ $(0.024)$& ${\bf 0.036}$ $({\bf 0.001})$  \\ 
3& $0.218$ $(0.016)$& $0.231$ $(0.020)$& ${\bf 0.213}$ $({\bf 0.016})$  & 12& ${\bf 0.761}$ $({\bf 0.075})$& $0.787$ $(0.091)$& $0.792$ $(0.100)$  \\ 
4& ${\bf 0.558}$ $({\bf 0.027})$& $0.567$ $(0.029)$& $0.561$ $(0.030)$  & 13& $0.109$ $(0.017)$& $0.109$ $(0.017)$& ${\bf 0.068}$ $({\bf 0.010})$  \\ 
5& $0.058$ $(0.004)$& $0.060$ $(0.005)$& ${\bf 0.055}$ $({\bf 0.003})$  & 14& $0.019$ $(0.001)$& $0.019$ $(0.001)$& ${\bf 0.019}$ $({\bf 0.000})$  \\ 
6& $0.567$ $(0.081)$& $0.634$ $(0.103)$& ${\bf 0.459}$ $({\bf 0.045})$  & 15& $0.076$ $(0.008)$& $0.078$ $(0.009)$& ${\bf 0.071}$ $({\bf 0.006})$  \\ 
7& $0.020$ $(0.012)$& $0.022$ $(0.014)$& ${\bf 0.019}$ $({\bf 0.005})$  & 16& $0.172$ $(0.032)$& $0.192$ $(0.036)$& ${\bf 0.119}$ $({\bf 0.014})$  \\ 
8& $0.395$ $(0.395)$& $0.463$ $(0.414)$& ${\bf 0.263}$ $({\bf 0.352})$  & 17& $0.041$ $(0.004)$& $0.043$ $(0.005)$& ${\bf 0.040}$ $({\bf 0.004})$  \\ 
9& $0.437$ $(0.096)$& $0.367$ $(0.060)$& ${\bf 0.222}$ $({\bf 0.015})$  & 18& $0.036$ $(0.007)$& $0.039$ $(0.007)$& ${\bf 0.028}$ $({\bf 0.009})$  \\ 
\hline
\multicolumn{8}{l}{$n=400$}\\
\hline
1& $0.051$ $(0.003)$& $0.052$ $(0.003)$& ${\bf 0.050}$ $({\bf 0.002})$  & 10& $0.218$ $(0.022)$& $0.233$ $(0.027)$& ${\bf 0.192}$ $({\bf 0.010})$  \\ 
2& $0.177$ $(0.008)$& $0.172$ $(0.006)$& ${\bf 0.167}$ $({\bf 0.005})$  & 11& $0.051$ $(0.009)$& $0.122$ $(0.031)$& ${\bf 0.036}$ $({\bf 0.001})$  \\ 
3& $0.199$ $(0.011)$& $0.209$ $(0.013)$& ${\bf 0.193}$ $({\bf 0.010})$  & 12& ${\bf 0.691}$ $({\bf 0.040})$& $0.701$ $(0.051)$& $0.709$ $(0.058)$  \\ 
4& $0.517$ $(0.018)$& $0.527$ $(0.019)$& ${\bf 0.510}$ $({\bf 0.016})$  & 13& $0.070$ $(0.009)$& $0.072$ $(0.008)$& ${\bf 0.054}$ $({\bf 0.004})$  \\ 
5& $0.050$ $(0.003)$& $0.051$ $(0.003)$& ${\bf 0.050}$ $({\bf 0.002})$  & 14& $0.019$ $(0.001)$& $0.019$ $(0.001)$& ${\bf 0.019}$ $({\bf 0.000})$  \\ 
6& $0.513$ $(0.055)$& $0.555$ $(0.063)$& ${\bf 0.432}$ $({\bf 0.036})$  & 15& $0.059$ $(0.004)$& $0.060$ $(0.005)$& ${\bf 0.057}$ $({\bf 0.003})$  \\ 
7& $0.019$ $(0.010)$& $0.021$ $(0.012)$& ${\bf 0.014}$ $({\bf 0.003})$  & 16& $0.105$ $(0.014)$& $0.106$ $(0.014)$& ${\bf 0.090}$ $({\bf 0.007})$  \\ 
8& $0.209$ $(0.171)$& $0.286$ $(0.248)$& ${\bf 0.110}$ $({\bf 0.107})$  & 17& ${\bf 0.032}$ $({\bf 0.002})$& $0.033$ $(0.003)$& ${\bf 0.032}$ $({\bf 0.002})$  \\ 
9& $0.249$ $(0.024)$& $0.304$ $(0.037)$& ${\bf 0.201}$ $({\bf 0.013})$  & 18& $0.029$ $(0.006)$& $0.032$ $(0.005)$& ${\bf 0.023}$ $({\bf 0.006})$  \\ 
\hline
\end{tabular}
\end{center}
\vspace{-10pt}
\end{table}

\section{Conclusion}

%Recently, random features have been shown to be a powerful tool to cheaply learn nonlinear functions. Furthermore, \texttt{SSSL} based on Nystr\"om random features has been shown to speed up learning when few labeled training points are available but there are many unlabelled points. 

We have introduced the \cksny algorithm for semi-supervised learning. By combining two randomly generated views of Nystr\"om features via an efficient implementation of CCA, \cksny outperforms the prior state-of-the-art, \sssl, by  10-15\% (depending on the number of labeled points) on average over 18 datasets. Furthermore, \cksny is over 3 orders of magnitude faster than \sssl on medium sized datasets ($N=10,000$) with further gains as $N$ increases. An interesting research direction is to investigate using the recently developed deep CCA algorithm, which extracts higher order correlations between views \cite{andrew:13}, as a preprocessing step.

In this work we use a uniform sampling scheme for the Nystr\"om method for computational reasons since it has been shown to perform well empirically relative to more expensive schemes \cite{kumar:12}. Since CCA gives us a criterion by which to measure the important of random features, in the future we aim to investigate active sampling schemes based on canonical correlations which may yield better performance by selecting the most informative indices to sample.

\paragraph{Acknowledgements.}
We thank Haim Avron for help with implementing randomized CCA and Patrick Pletscher for drawing our attention to the Nystr\"om method.

\FloatBarrier
{\small
\bibliography{cks}
\bibliographystyle{bmc_article}
}

\newpage
\FloatBarrier
{\Large{\textbf{Supplementary Information}}}
\newcounter{si-sec}
\renewcommand{\thesection}{SI.\arabic{si-sec}}

\FloatBarrier
\addtocounter{si-sec}{1}
\section{Complete \cksny results}
\label{s:xnvres}

\begin{table}[htp]
\begin{center}
\caption{Performance (normalized MSE/classification error rate). Standard errors in parentheses. \label{tab:results2}}
\setlength{\tabcolsep}{3.0pt}
\small
\begin{tabular}{l r r r | l r r r }
\hline
\multicolumn{1}{c}{set}&
\multicolumn{1}{c}{\ssslm}&
\multicolumn{1}{c}{\ssslM}&
\multicolumn{1}{c}{\cksny}
 & set &
\multicolumn{1}{c}{\ssslm}&
\multicolumn{1}{c}{\ssslM}&
\multicolumn{1}{c}{\cksny} \\ 
\hline 
\multicolumn{8}{l}{$n=100$}\\
%$n=$\\100\\
\hline
1& ${\bf 0.058}$ $({\bf 0.008})$& $0.060$ $(0.009)$& $0.059$ $(0.008)$  & 10& $0.439$ $(0.088)$& $0.545$ $(0.121)$& ${\bf 0.286}$ $({\bf 0.040})$  \\ 
2& $0.220$ $(0.015)$& $0.200$ $(0.016)$& ${\bf 0.184}$ $({\bf 0.016})$  & 11& $0.064$ $(0.025)$& $0.054$ $(0.015)$& ${\bf 0.037}$ $({\bf 0.001})$  \\ 
3& ${\bf 0.249}$ $({\bf 0.024})$& $0.263$ $(0.028)$& $0.255$ $(0.029)$  & 12& ${\bf 0.825}$ $({\bf 0.114})$& $0.864$ $(0.144)$& $0.895$ $(0.163)$  \\ 
4& ${\bf 0.651}$ $({\bf 0.063})$& $0.666$ $(0.070)$& $0.691$ $(0.082)$  & 13& $0.160$ $(0.026)$& $0.167$ $(0.027)$& ${\bf 0.104}$ $({\bf 0.024})$  \\ 
5& $0.068$ $(0.008)$& $0.076$ $(0.012)$& ${\bf 0.061}$ $({\bf 0.005})$  & 14& $0.020$ $(0.003)$& $0.020$ $(0.003)$& ${\bf 0.019}$ $({\bf 0.000})$  \\ 
6& $0.628$ $(0.122)$& $0.718$ $(0.153)$& ${\bf 0.504}$ $({\bf 0.074})$  & 15& $0.104$ $(0.015)$& $0.104$ $(0.016)$& ${\bf 0.095}$ $({\bf 0.013})$  \\ 
7& ${\bf 0.029}$ $({\bf 0.016})$& $0.031$ $(0.019)$& $0.036$ $(0.020)$  & 16& $0.231$ $(0.047)$& $0.261$ $(0.057)$& ${\bf 0.163}$ $({\bf 0.026})$  \\ 
8& $0.691$ $(0.603)$& $0.751$ $(0.659)$& ${\bf 0.568}$ $({\bf 0.613})$  & 17& $0.058$ $(0.010)$& $0.061$ $(0.011)$& ${\bf 0.056}$ $({\bf 0.009})$  \\ 
9& $0.488$ $(0.123)$& $0.367$ $(0.073)$& ${\bf 0.276}$ $({\bf 0.047})$  & 18& $0.042$ $(0.009)$& $0.043$ $(0.009)$& ${\bf 0.036}$ $({\bf 0.011})$  \\ 
\hline
\multicolumn{8}{l}{$n=200$}\\
\hline
1& $0.054$ $(0.005)$& $0.055$ $(0.006)$& ${\bf 0.053}$ $({\bf 0.004})$  & 10& $0.309$ $(0.059)$& $0.358$ $(0.077)$& ${\bf 0.226}$ $({\bf 0.020})$  \\ 
2& $0.198$ $(0.014)$& $0.184$ $(0.010)$& ${\bf 0.175}$ $({\bf 0.010})$  & 11& $0.146$ $(0.048)$& $0.072$ $(0.024)$& ${\bf 0.036}$ $({\bf 0.001})$  \\ 
3& $0.218$ $(0.016)$& $0.231$ $(0.020)$& ${\bf 0.213}$ $({\bf 0.016})$  & 12& ${\bf 0.761}$ $({\bf 0.075})$& $0.787$ $(0.091)$& $0.792$ $(0.100)$  \\ 
4& ${\bf 0.558}$ $({\bf 0.027})$& $0.567$ $(0.029)$& $0.561$ $(0.030)$  & 13& $0.109$ $(0.017)$& $0.109$ $(0.017)$& ${\bf 0.068}$ $({\bf 0.010})$  \\ 
5& $0.058$ $(0.004)$& $0.060$ $(0.005)$& ${\bf 0.055}$ $({\bf 0.003})$  & 14& $0.019$ $(0.001)$& $0.019$ $(0.001)$& ${\bf 0.019}$ $({\bf 0.000})$  \\ 
6& $0.567$ $(0.081)$& $0.634$ $(0.103)$& ${\bf 0.459}$ $({\bf 0.045})$  & 15& $0.076$ $(0.008)$& $0.078$ $(0.009)$& ${\bf 0.071}$ $({\bf 0.006})$  \\ 
7& $0.020$ $(0.012)$& $0.022$ $(0.014)$& ${\bf 0.019}$ $({\bf 0.005})$  & 16& $0.172$ $(0.032)$& $0.192$ $(0.036)$& ${\bf 0.119}$ $({\bf 0.014})$  \\ 
8& $0.395$ $(0.395)$& $0.463$ $(0.414)$& ${\bf 0.263}$ $({\bf 0.352})$  & 17& $0.041$ $(0.004)$& $0.043$ $(0.005)$& ${\bf 0.040}$ $({\bf 0.004})$  \\ 
9& $0.437$ $(0.096)$& $0.367$ $(0.060)$& ${\bf 0.222}$ $({\bf 0.015})$  & 18& $0.036$ $(0.007)$& $0.039$ $(0.007)$& ${\bf 0.028}$ $({\bf 0.009})$  \\ 
\hline
\multicolumn{8}{l}{$n=300$}\\
\hline
1& $0.052$ $(0.004)$& $0.053$ $(0.004)$& ${\bf 0.051}$ $({\bf 0.003})$ & 10& $0.250$ $(0.031)$& $0.275$ $(0.040)$& ${\bf 0.205}$ $({\bf 0.014})$ \\
2& $0.185$ $(0.011)$& $0.177$ $(0.008)$& ${\bf 0.171}$ $({\bf 0.007})$ & 11& $0.074$ $(0.020)$& $0.105$ $(0.032)$& ${\bf 0.036}$ $({\bf 0.001})$ \\
3& $0.206$ $(0.012)$& $0.217$ $(0.015)$& ${\bf 0.200}$ $({\bf 0.012})$  & 12& ${\bf 0.719}$ $({\bf 0.052})$& $0.736$ $(0.067)$& $0.744$ $(0.083)$ \\
4& $0.531$ $(0.020)$& $0.540$ $(0.021)$& ${\bf 0.526}$ $({\bf 0.020})$ & 13& $0.083$ $(0.010)$& $0.084$ $(0.009)$& ${\bf 0.058}$ $({\bf 0.006})$ \\
5& $0.053$ $(0.004)$& $0.055$ $(0.004)$& ${\bf 0.052}$ $({\bf 0.003})$ & 14& $0.019$ $(0.002)$& $0.019$ $(0.002)$& ${\bf 0.019}$ $({\bf 0.000})$ \\
6& $0.535$ $(0.065)$& $0.585$ $(0.079)$& ${\bf 0.444}$ $({\bf 0.039})$ & 15& $0.066$ $(0.005)$& $0.067$ $(0.006)$& ${\bf 0.062}$ $({\bf 0.004})$  \\
7& $0.020$ $(0.010)$& $0.022$ $(0.013)$& ${\bf 0.016}$ $({\bf 0.003})$ & 16& $0.126$ $(0.020)$& $0.133$ $(0.022)$& ${\bf 0.100}$ $({\bf 0.009})$ \\
8& $0.270$ $(0.216)$& $0.370$ $(0.333)$& ${\bf 0.152}$ $({\bf 0.199})$ & 17& $0.035$ $(0.003)$& $0.037$ $(0.004)$& ${\bf 0.035}$ $({\bf 0.002})$ \\
9& $0.304$ $(0.038)$& $0.352$ $(0.055)$& ${\bf 0.207}$ $({\bf 0.013})$ & 18& $0.032$ $(0.006)$& $0.035$ $(0.007)$& ${\bf 0.025}$ $({\bf 0.006})$ \\
\hline
\multicolumn{8}{l}{$n=400$}\\
\hline
1& $0.051$ $(0.003)$& $0.052$ $(0.003)$& ${\bf 0.050}$ $({\bf 0.002})$  & 10& $0.218$ $(0.022)$& $0.233$ $(0.027)$& ${\bf 0.192}$ $({\bf 0.010})$  \\ 
2& $0.177$ $(0.008)$& $0.172$ $(0.006)$& ${\bf 0.167}$ $({\bf 0.005})$  & 11& $0.051$ $(0.009)$& $0.122$ $(0.031)$& ${\bf 0.036}$ $({\bf 0.001})$  \\ 
3& $0.199$ $(0.011)$& $0.209$ $(0.013)$& ${\bf 0.193}$ $({\bf 0.010})$  & 12& ${\bf 0.691}$ $({\bf 0.040})$& $0.701$ $(0.051)$& $0.709$ $(0.058)$  \\ 
4& $0.517$ $(0.018)$& $0.527$ $(0.019)$& ${\bf 0.510}$ $({\bf 0.016})$  & 13& $0.070$ $(0.009)$& $0.072$ $(0.008)$& ${\bf 0.054}$ $({\bf 0.004})$  \\ 
5& $0.050$ $(0.003)$& $0.051$ $(0.003)$& ${\bf 0.050}$ $({\bf 0.002})$  & 14& $0.019$ $(0.001)$& $0.019$ $(0.001)$& ${\bf 0.019}$ $({\bf 0.000})$  \\ 
6& $0.513$ $(0.055)$& $0.555$ $(0.063)$& ${\bf 0.432}$ $({\bf 0.036})$  & 15& $0.059$ $(0.004)$& $0.060$ $(0.005)$& ${\bf 0.057}$ $({\bf 0.003})$  \\ 
7& $0.019$ $(0.010)$& $0.021$ $(0.012)$& ${\bf 0.014}$ $({\bf 0.003})$  & 16& $0.105$ $(0.014)$& $0.106$ $(0.014)$& ${\bf 0.090}$ $({\bf 0.007})$  \\ 
8& $0.209$ $(0.171)$& $0.286$ $(0.248)$& ${\bf 0.110}$ $({\bf 0.107})$  & 17& ${\bf 0.032}$ $({\bf 0.002})$& $0.033$ $(0.003)$& ${\bf 0.032}$ $({\bf 0.002})$  \\ 
9& $0.249$ $(0.024)$& $0.304$ $(0.037)$& ${\bf 0.201}$ $({\bf 0.013})$  & 18& $0.029$ $(0.006)$& $0.032$ $(0.005)$& ${\bf 0.023}$ $({\bf 0.006})$  \\ 
\hline
\multicolumn{8}{l}{$n=500$}\\
\hline
1& $0.051$ $(0.002)$& $0.051$ $(0.003)$& ${\bf 0.050}$ $({\bf 0.002})$  & 10& $0.202$ $(0.017)$& $0.214$ $(0.020)$& ${\bf 0.185}$ $({\bf 0.008})$  \\ 
2& $0.172$ $(0.007)$& $0.169$ $(0.005)$& ${\bf 0.165}$ $({\bf 0.004})$  & 11& $0.043$ $(0.005)$& $0.092$ $(0.018)$& ${\bf 0.036}$ $({\bf 0.001})$  \\ 
3& $0.194$ $(0.008)$& $0.202$ $(0.010)$& ${\bf 0.188}$ $({\bf 0.007})$  & 12& ${\bf 0.675}$ $({\bf 0.035})$& $0.680$ $(0.044)$& $0.686$ $(0.047)$  \\ 
4& $0.508$ $(0.012)$& $0.517$ $(0.014)$& ${\bf 0.499}$ $({\bf 0.011})$  & 13& $0.061$ $(0.006)$& $0.063$ $(0.006)$& ${\bf 0.051}$ $({\bf 0.004})$  \\ 
5& $0.048$ $(0.002)$& $0.049$ $(0.002)$& ${\bf 0.048}$ $({\bf 0.002})$  & 14& ${\bf 0.019}$ $({\bf 0.000})$& ${\bf 0.019}$ $({\bf 0.000})$& ${\bf 0.019}$ $({\bf 0.000})$  \\ 
6& $0.503$ $(0.052)$& $0.541$ $(0.060)$& ${\bf 0.427}$ $({\bf 0.034})$  & 15& $0.055$ $(0.003)$& $0.055$ $(0.004)$& ${\bf 0.054}$ $({\bf 0.002})$  \\ 
7& $0.017$ $(0.007)$& $0.018$ $(0.007)$& ${\bf 0.014}$ $({\bf 0.002})$  & 16& $0.089$ $(0.010)$& $0.088$ $(0.010)$& ${\bf 0.083}$ $({\bf 0.005})$  \\ 
8& $0.167$ $(0.137)$& $0.241$ $(0.235)$& ${\bf 0.098}$ $({\bf 0.097})$  & 17& ${\bf 0.030}$ $({\bf 0.002})$& $0.030$ $(0.002)$& $0.031$ $(0.001)$  \\ 
9& $0.222$ $(0.017)$& $0.259$ $(0.027)$& ${\bf 0.196}$ $({\bf 0.011})$  & 18& $0.027$ $(0.004)$& $0.029$ $(0.005)$& ${\bf 0.022}$ $({\bf 0.005})$  \\ 
\hline
\end{tabular}
\end{center}
\end{table}

%%%%%%%%%%%%%%%
% Figures
%%%%%%%%%%%%%%%

%%%%%%% prediction error plots set 2
\begin{figure}
\begin{centering}
\vspace{-10pt}
\subfloat[\texttt{abalone}]{\includegraphics[width=0.3\columnwidth]{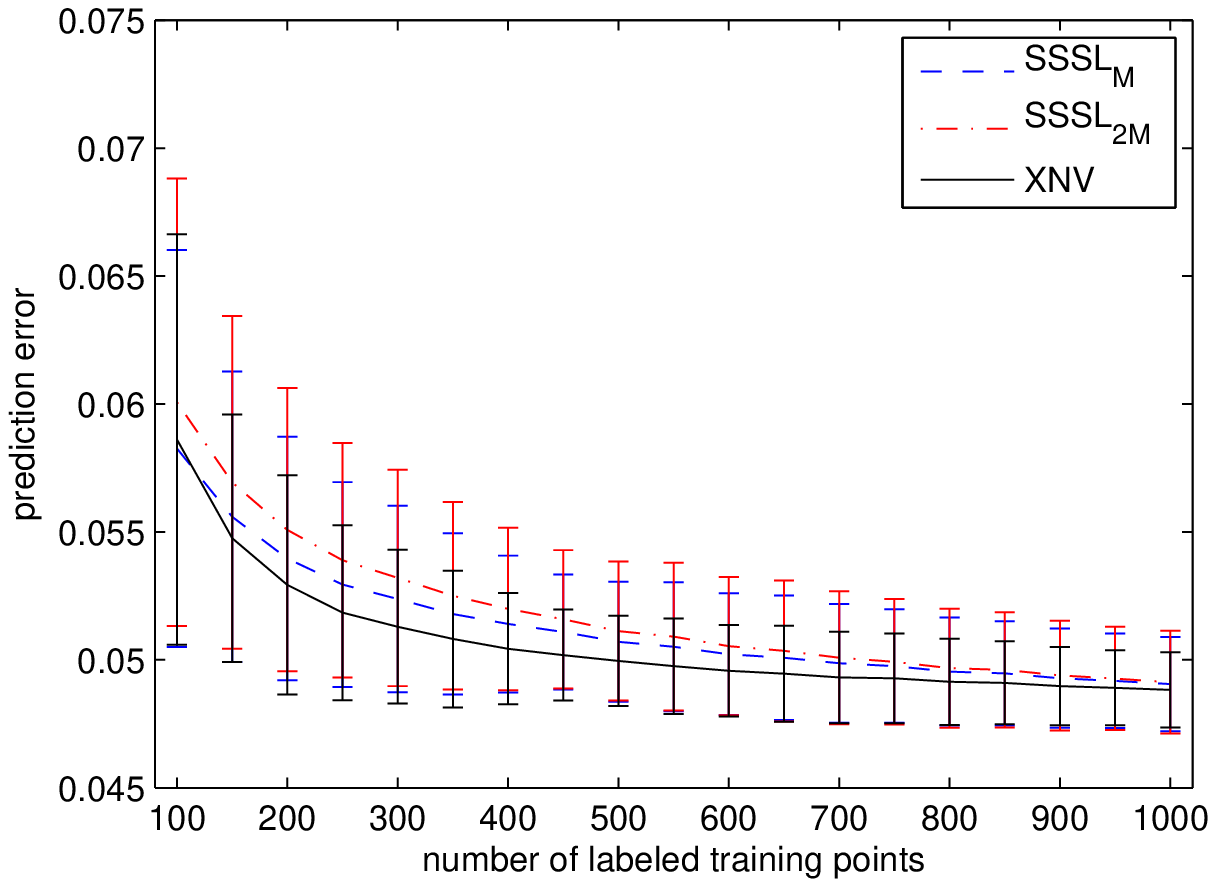} \label{fig:abalone}} 
\subfloat[\texttt{adult}]{\includegraphics[width=0.3\columnwidth]{ny/eps/19_adult} \label{fig:adult}} 
\subfloat[\texttt{ailerons}]{\includegraphics[width=0.3\columnwidth]{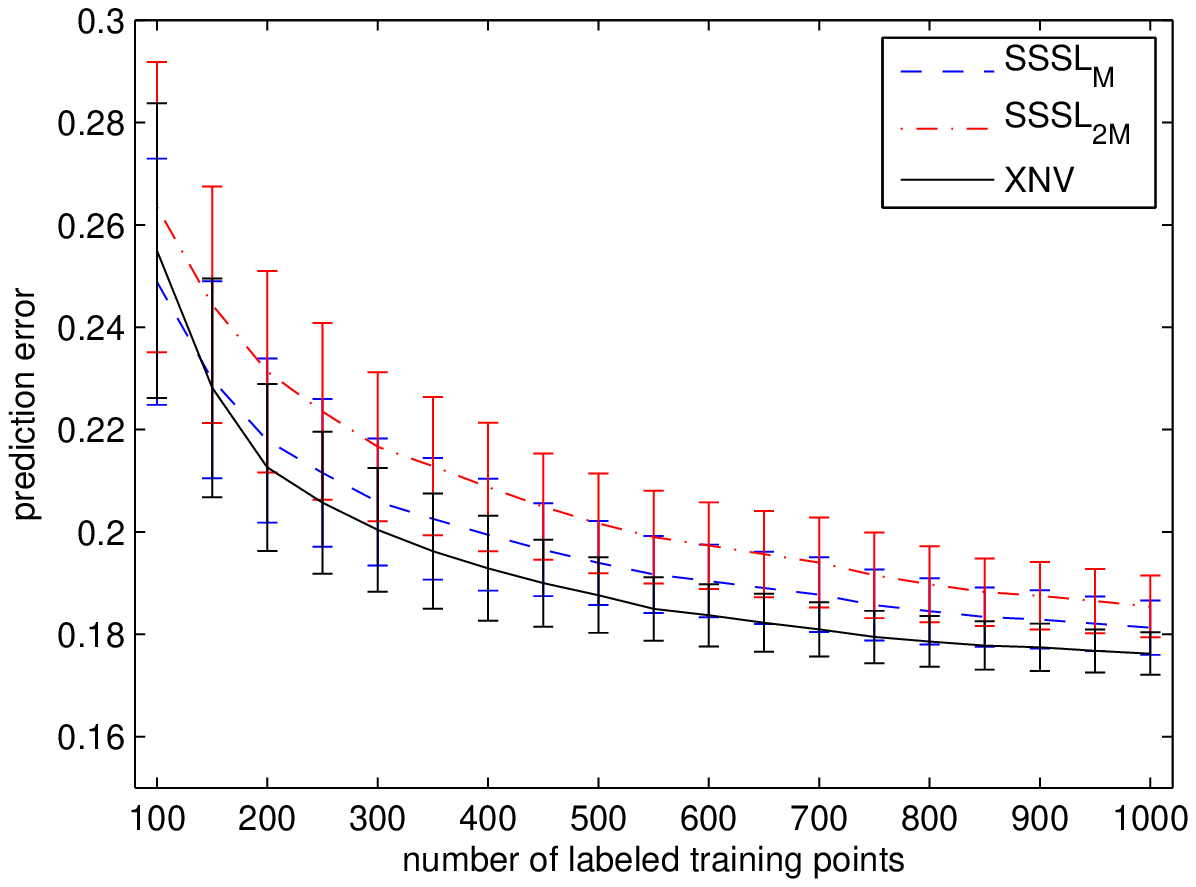} \label{fig:ailerons}} 
\vspace{-10pt}
\subfloat[\texttt{bank8}]{\includegraphics[width=0.3\columnwidth]{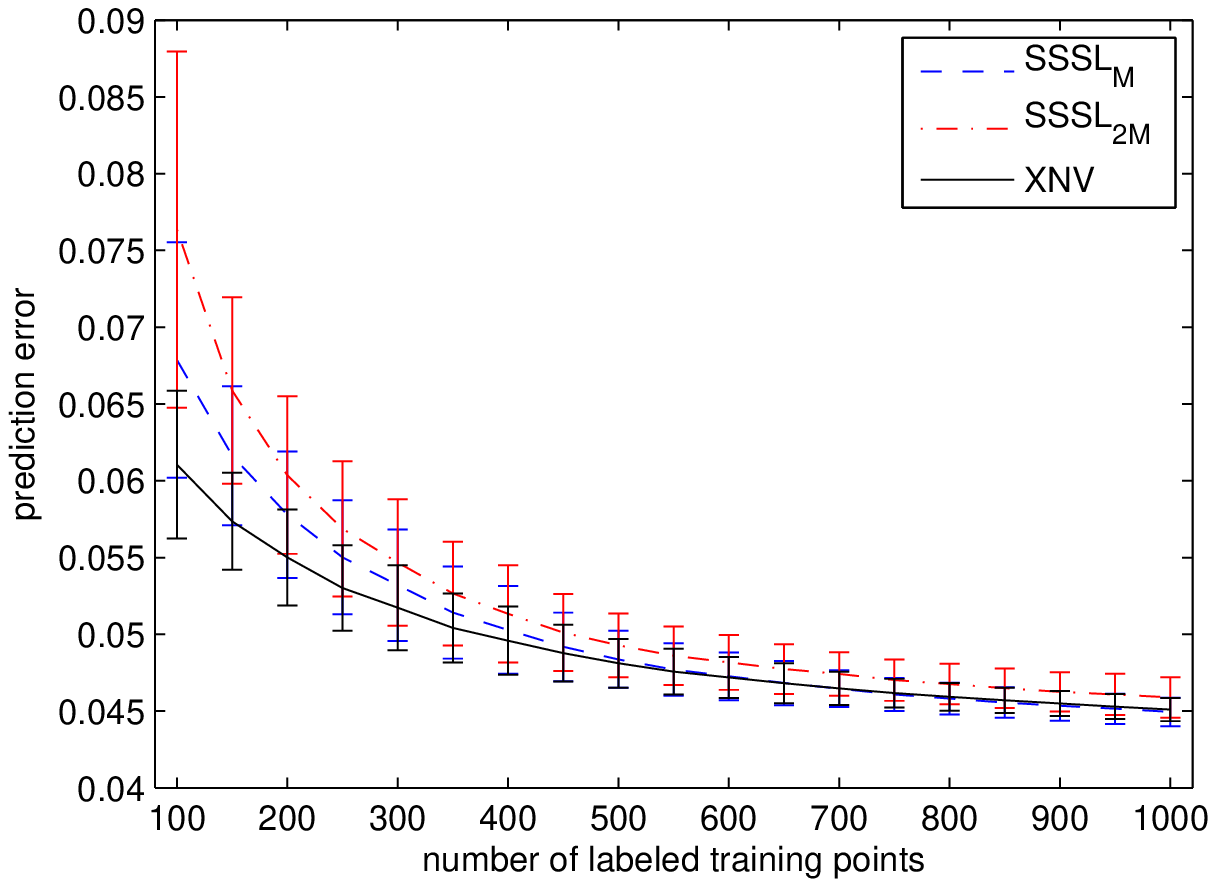} \label{fig:bank8}} 
\subfloat[\texttt{bank32}]{\includegraphics[width=0.3\columnwidth]{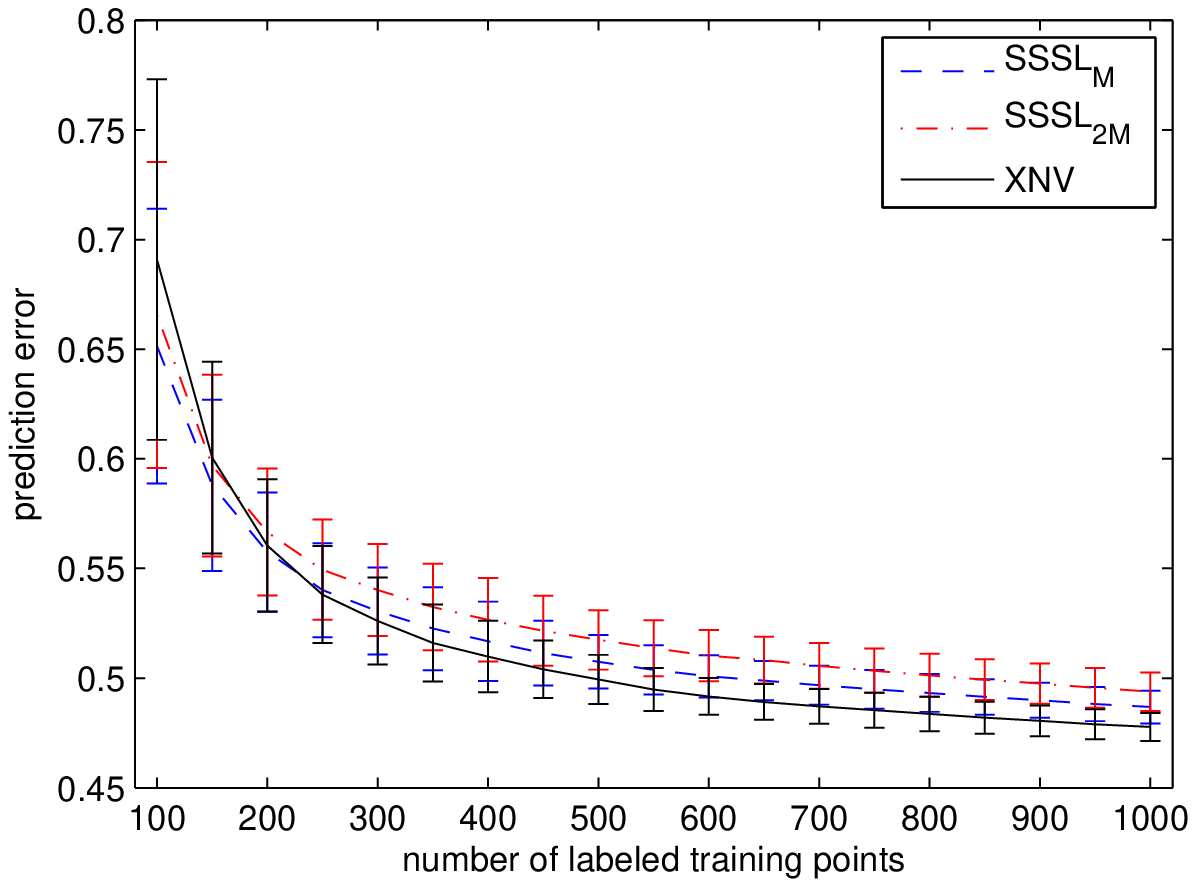} \label{fig:bank32}} 
\subfloat[\texttt{cal housing}]{\includegraphics[width=0.3\columnwidth]{ny/eps/19_cal} \label{fig:cal}}
\vspace{-10pt}
\subfloat[ \texttt{census}]{\includegraphics[width=0.3\textwidth]{ny/eps/19_census} \label{fig:census}} 
\subfloat[ \texttt{CPU}]{\includegraphics[width=0.3\textwidth]{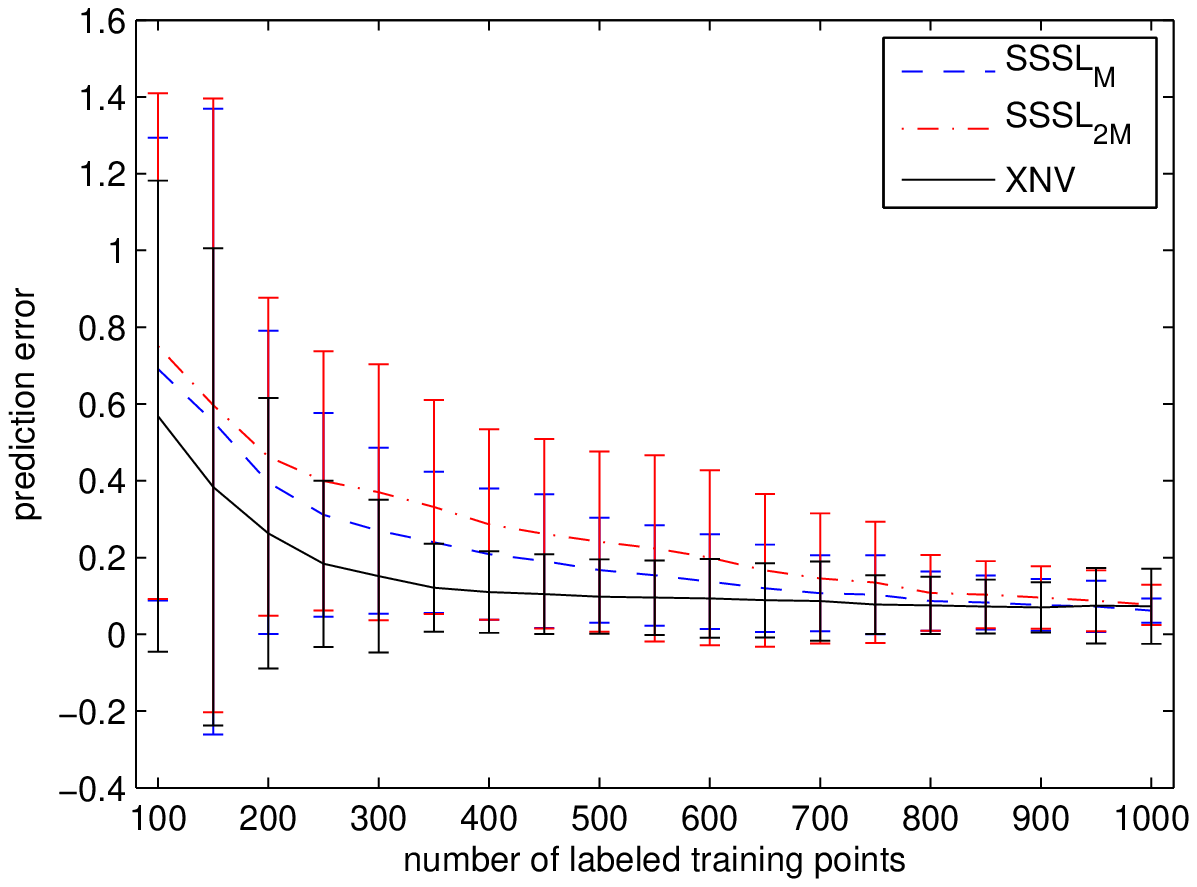} \label{fig:cpu}} 
\subfloat[ \texttt{CT}]{\includegraphics[width=0.3\textwidth]{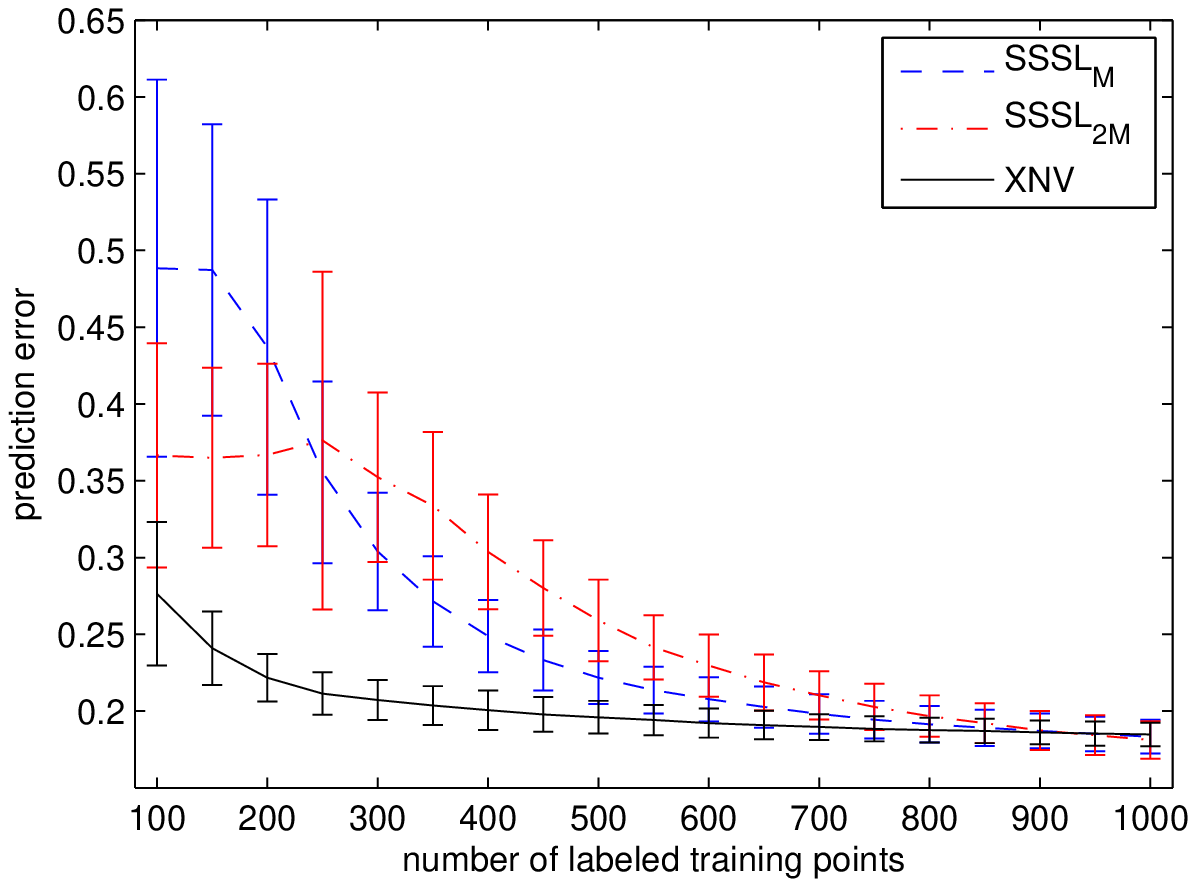} \label{fig:ct}}
\vspace{-10pt}
\subfloat[\texttt{elevators}]{\includegraphics[width=0.3\columnwidth]{ny/eps/19_elevators} \label{fig:elevators}} 
\subfloat[\texttt{HIVa}]{\includegraphics[width=0.3\columnwidth]{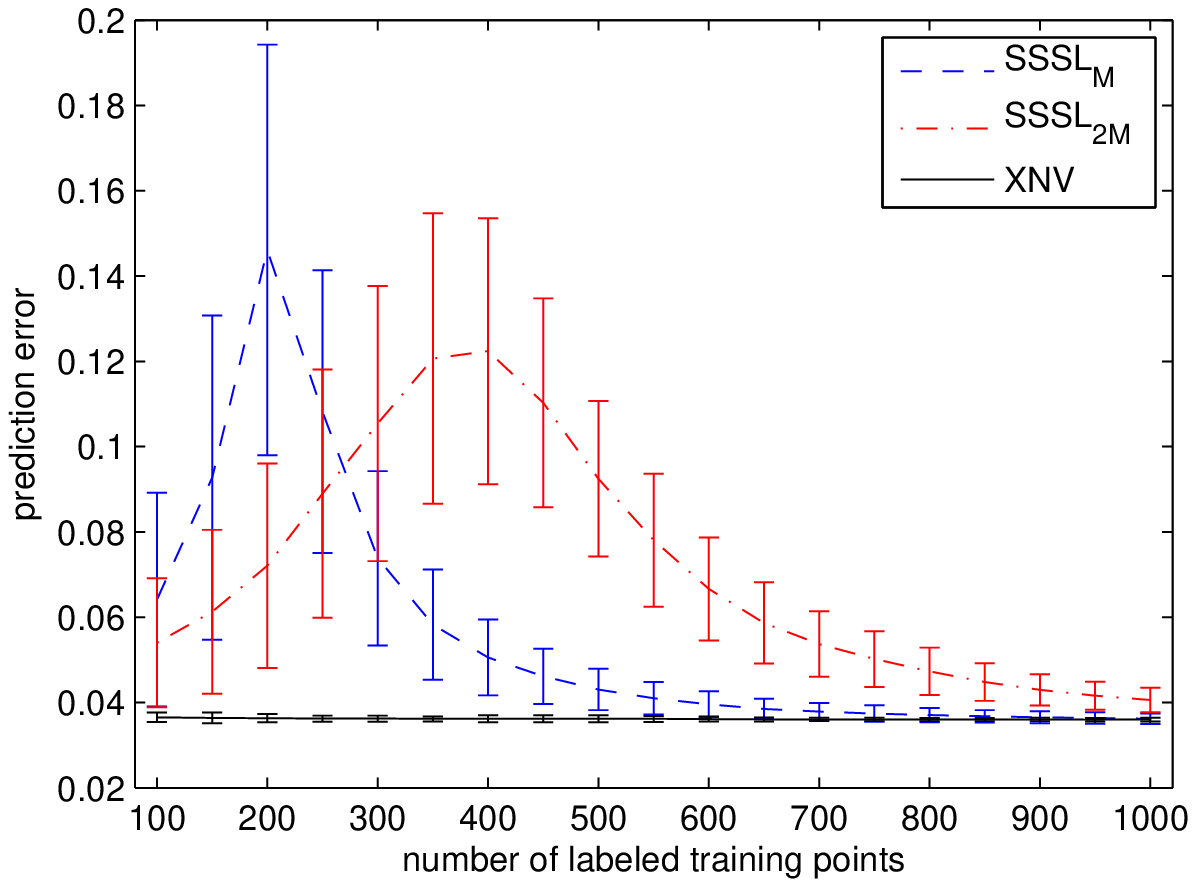} \label{fig:hiva}} 
\subfloat[\texttt{house}]{\includegraphics[width=0.3\columnwidth]{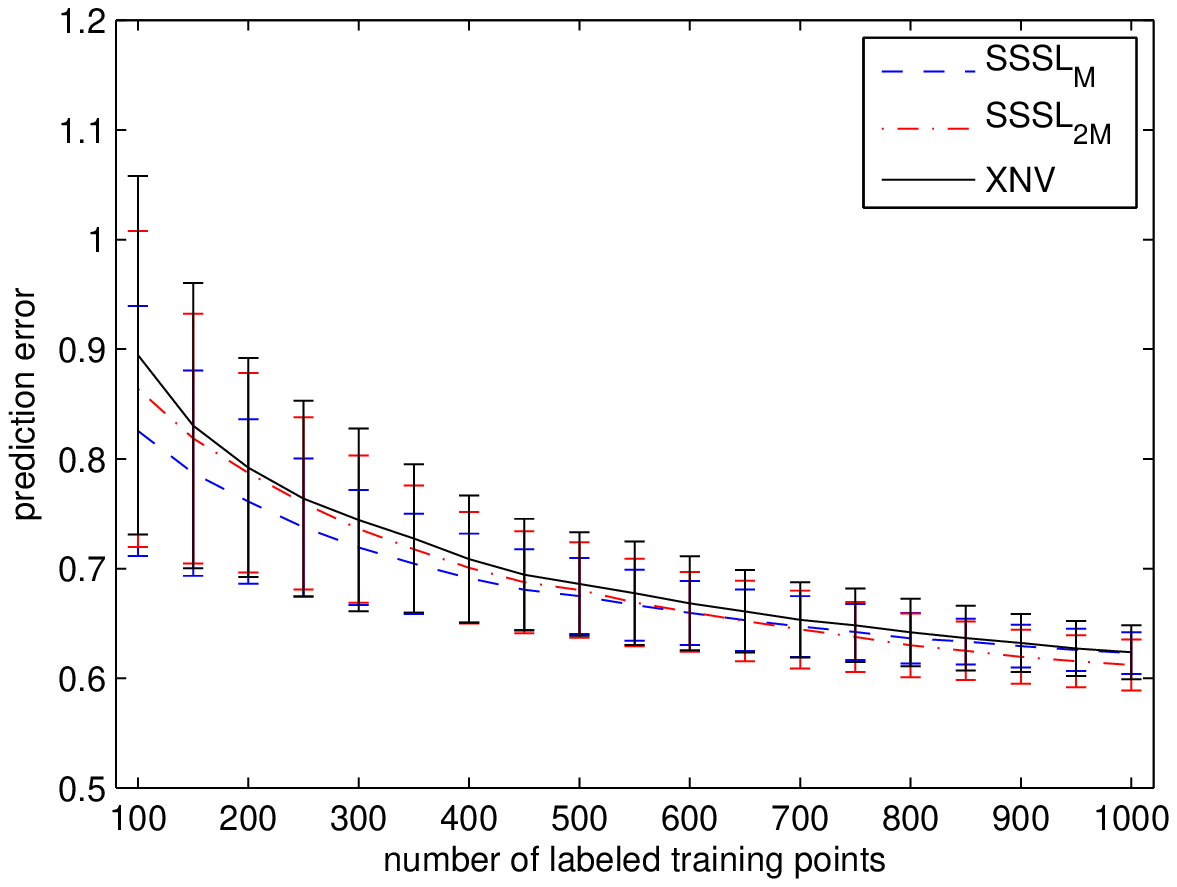} \label{fig:house}} 
\vspace{-10pt}
\subfloat[\texttt{ibn Sina}]{\includegraphics[width=0.3\columnwidth]{ny/eps/19_ibn_sina} \label{fig:ibnsina}} 
\subfloat[\texttt{orange}]{\includegraphics[width=0.3\columnwidth]{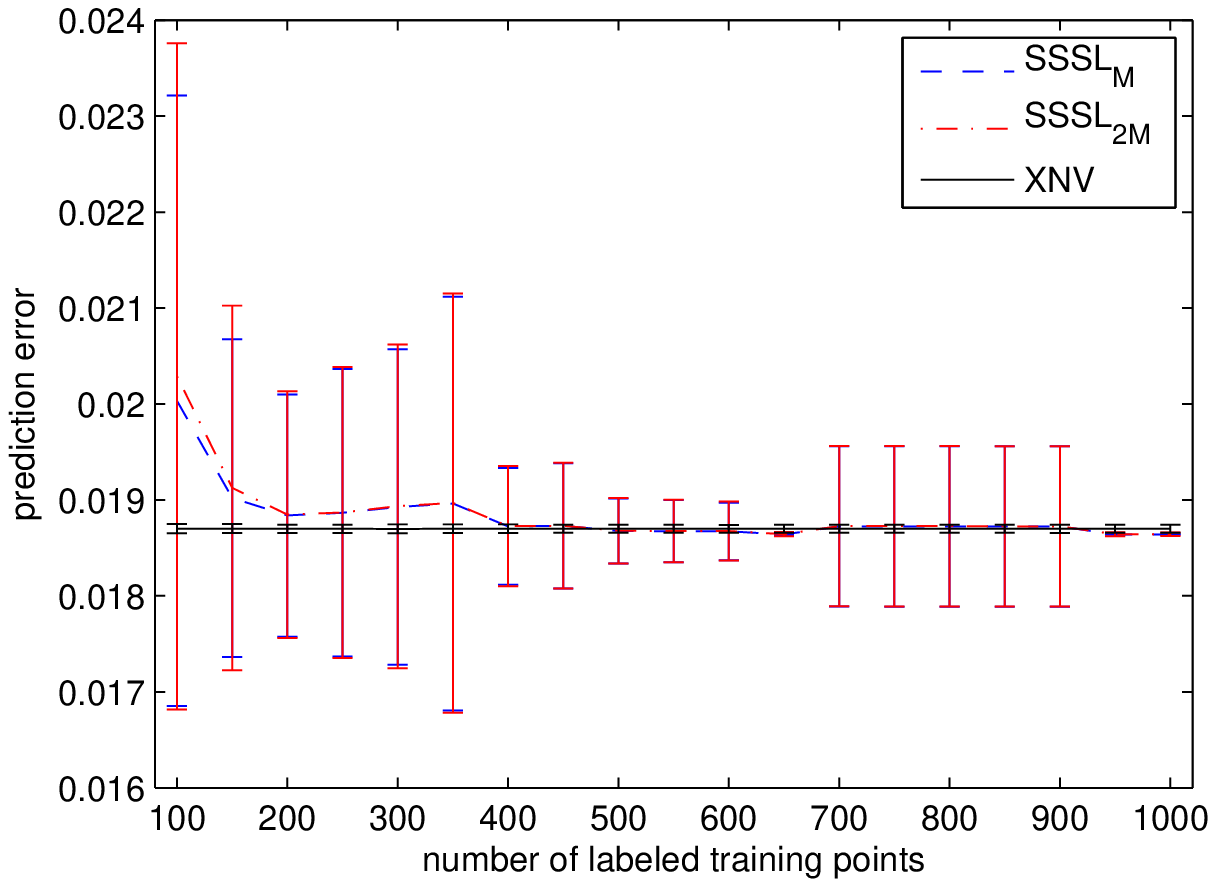} \label{fig:orange}} 
\subfloat[\texttt{sarcos 1}]{\includegraphics[width=0.3\columnwidth]{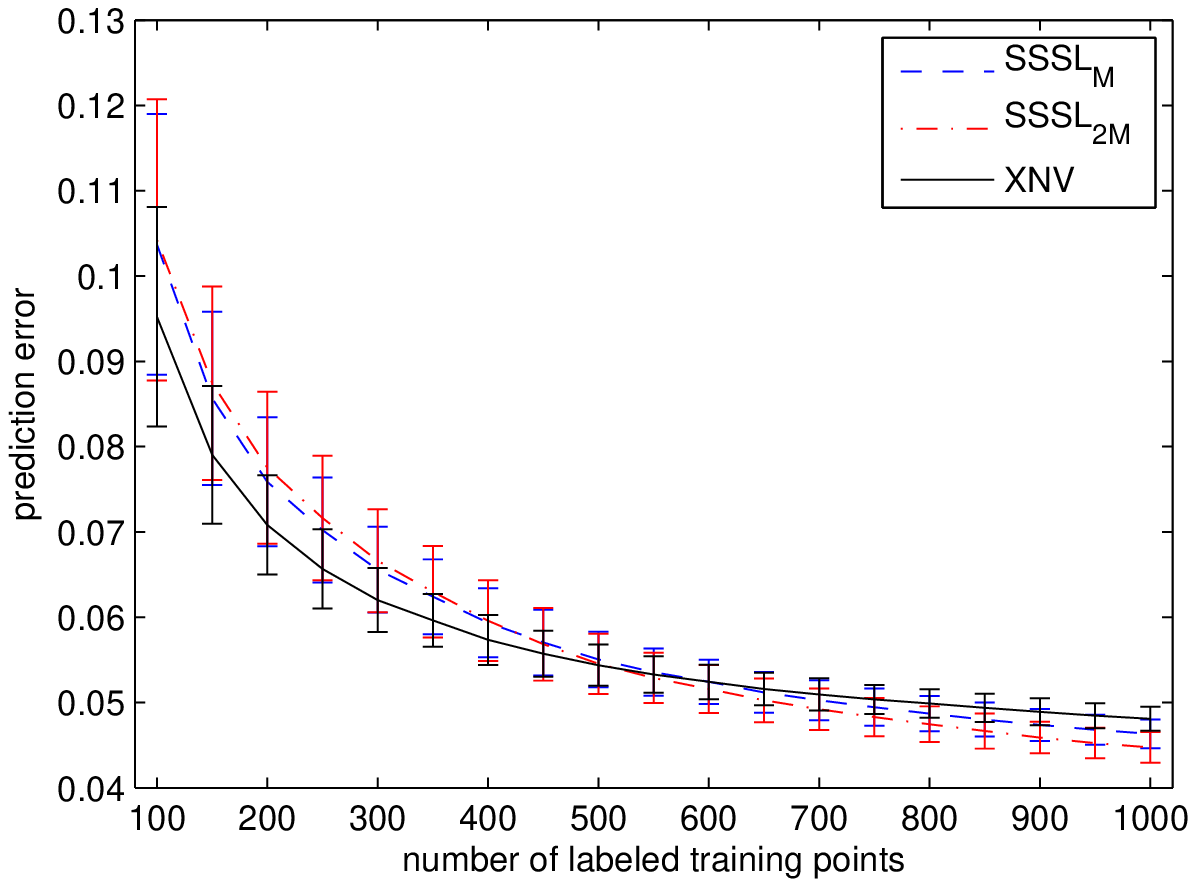} \label{fig:sarcos}}
\vspace{-10pt}
\subfloat[ \texttt{sarcos 5}]{\includegraphics[width=0.3\textwidth]{ny/eps/19_sarcos5} \label{fig:sarcos5}} 
\subfloat[ \texttt{sarcos 7}]{\includegraphics[width=0.3\textwidth]{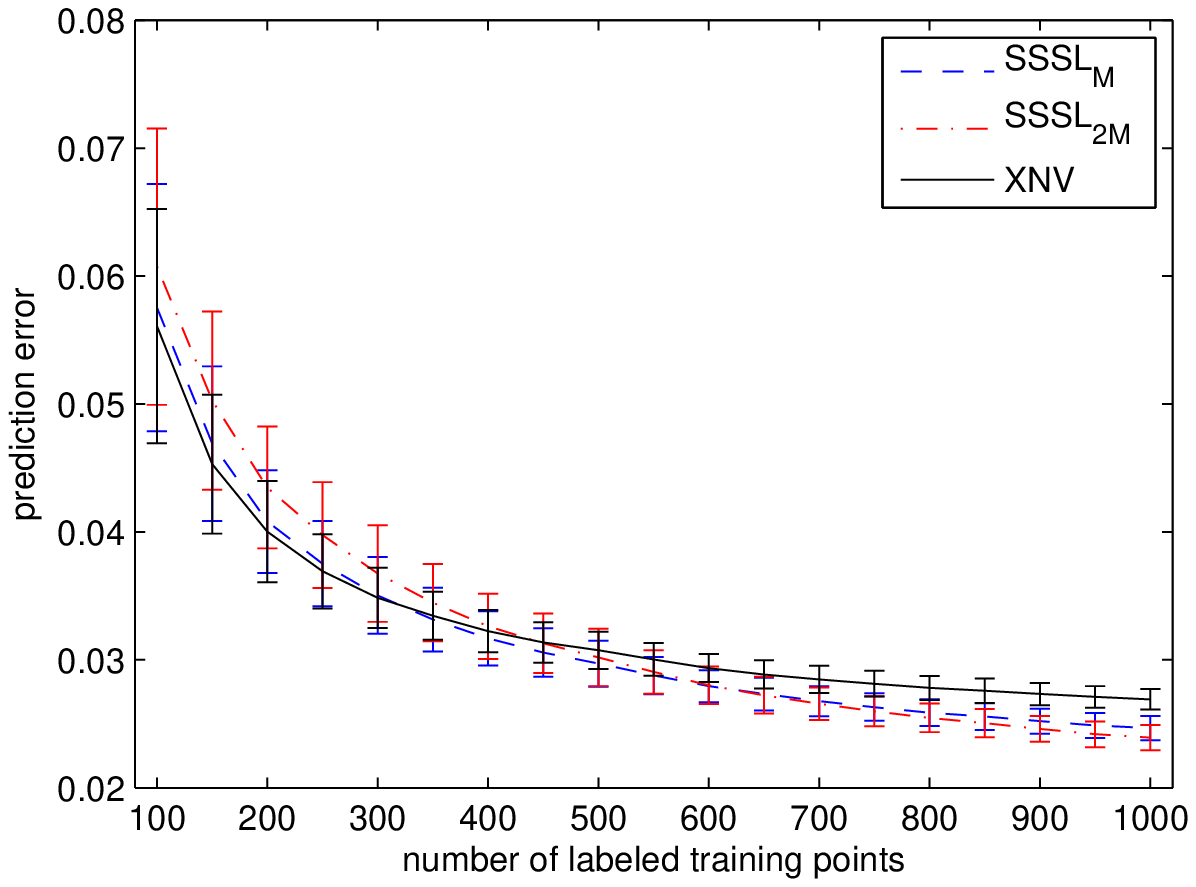} \label{fig:sarcos7}} 
\subfloat[ \texttt{sylva}]{\includegraphics[width=0.3\textwidth]{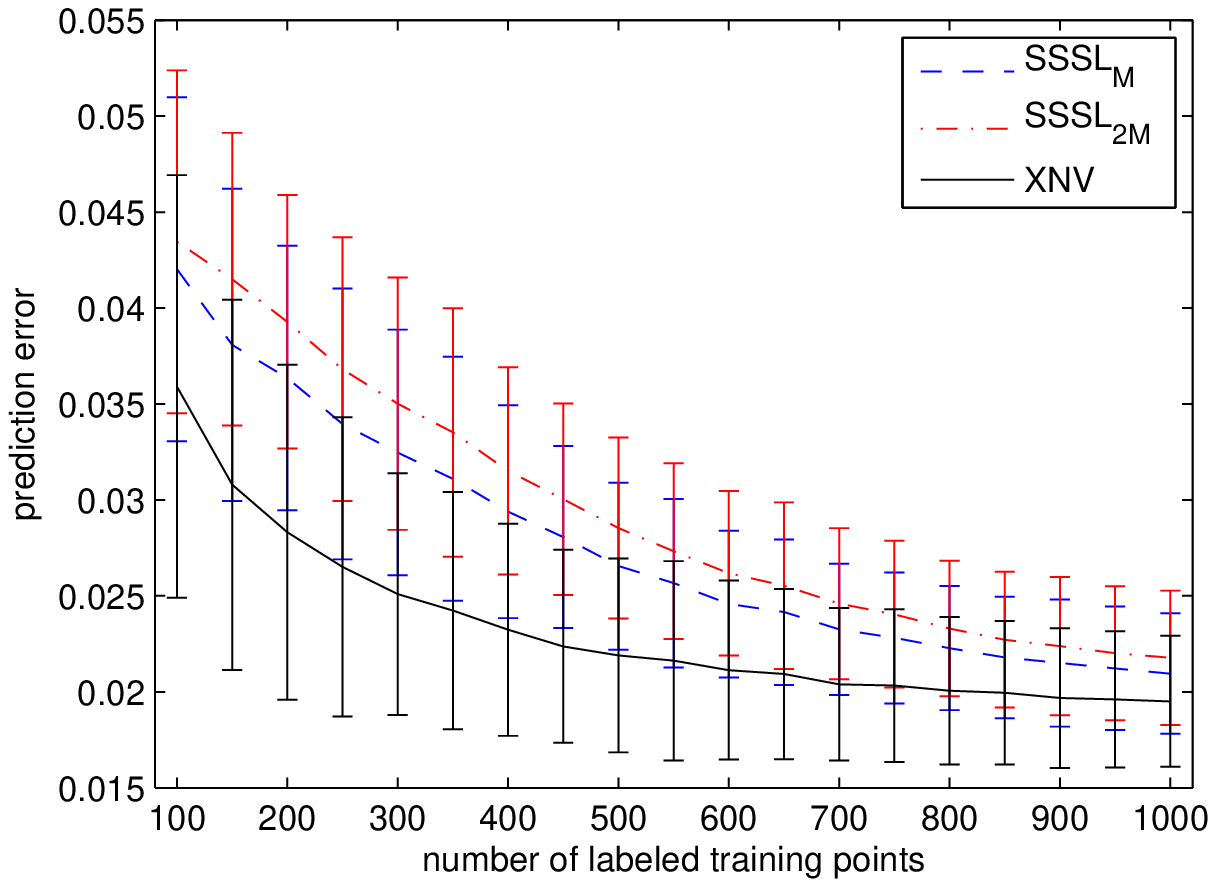} \label{fig:sylva}} 
\vspace{-5pt}
\caption{Comparison of mean prediction error and standard deviation on all 18 datasets. \label{fig:res3}}
\end{centering}
\end{figure}

%%%%%%%%%%%%%%%%%%%%%%%%%%%%%%%%%%%%%%%%%%%%%%%%%%%%%%%%%%%%%%%%%%%%%%%%%%%%%%
\addtocounter{si-sec}{1}
\section{Comparison with Kernel Ridge Regression}
\label{sec:krr}

We compare \ssslm and \cksny to kernel ridge regression (KRR). The table below reports the percentage improvement in mean error of both of these methods against KRR, averaged over the 18 datasets according to the experimental procedure detailed in \S\ref{sec:results}. Parameters $\sigma$ (kernel width) and $\gamma$ (ridge penalty) for KRR were chosen by 5-fold cross validation.  We observe that both \ssslm and \cksny far outperform KRR, by $50-60\%$. Importantly, this shows our approximation to \sssl far outperforms the fully supervised baseline.

\begin{center}
\small
\begin{tabular}{l r r r r r}
\hline
\ssslm and \cksny vs \texttt{KRR} & $n=100$ & $n=200$ & $n=300$ & $n=400$ & $n=500$ \\
\hline
Avg reduction in error for \ssslm  &  48\%  & 52\%  &  56\% &  58\% & 60\%  \\ 
Avg reduction in error for \cksny &  56\%  & 62\%  &  63\% &  63\% & 63\%  \\ 
\hline
\end{tabular}
\end{center}

%%%%%%%%%%%%%%%%%%%%%%%%%%%%%%%%%%%%%%%%%%%%%%%%%%%%%%%%%%%%%%%%%%%%%%%%%%%%%%
\addtocounter{si-sec}{1}
\section{Random Fourier features}
\label{s:rff}

Random Fourier features are a method for approximating shift invariant kernels \cite{rahimi:07}, i.e. where $\kappa(\x_i,\x_{i'}) = \kappa(\x_i - \x_{i'})$. Such a kernel function can be represented in terms of its inverse Fourier transform as $\kappa(\x_i - \x_{i'}) = \int_{\R^D} P(\boldomega) e^{j\boldomega\tr \br{\x_i - \x_{i'}} }$. $P(\boldomega)$ is the Fourier transform of $\kappa$ which is guaranteed to be a proper probability distribution and so for real-valued features $\kappa(\x_i,\x_{i'})$ can be equivalently interpreted as $\bE_{\boldomega} \sq{ \z(\x_i)\tr \z(\x_{i'})}$ where $\z(\x_i) = \frac{1}{\sqrt{2}} \cos(\boldomega\tr \x_i + b)$ . Replacing the expectation by the sample average leads to a scheme for constructing random features. In particular, a Gaussian kernel of width $\sigma$ has a Fourier transform which is also Gaussian. Sampling $\boldomega_m \sim \N(0,2\sigma\Id_D)$ and $b_m \sim \text{Unif}\sq{-\pi,\pi}$, we can then construct features whose inner product approximates this kernel as $\z_i = \frac{1}{\sqrt{M}}\sq{ \cos(\boldomega_1\tr \x_i + b_1),\ldots, \cos(\boldomega_M \tr \x_i + b_M) }$.

It was recently shown how both random Fourier features the Nystr\"om approximation could be cast in the same framework \cite{yang:12}. A major difference between the methods lies in the sampling scheme employed. Random Fourier features are constructed in a data independent fashion which makes them extremely cheap to compute. Nystr\"om features are constructed in a data dependent way which leads to improved performance but, in the case of semi-supervised learning, more expensive since we need to evaluate the approximate kernel for all unlabeled points we wish to use. %However, this is still orders of magnitude cheaper than constructing the full kernel as we shall see.

Algorithm \ref{alg:cks} details \texttt{Correlated Kitchen Sinks} (\cksrff). This algorithm generates random views using the random Fourier features procedure in step 1. Steps 2 and 3 proceed exactly as in Algorithm \ref{alg:xnv}.

\begin{algorithm}
	\caption{\texttt{Correlated Kitchen Sinks (\cksrff).} \label{alg:cks}}
	\algorithmicrequire\; Labeled data: $\{\x_i,y_i\}_{i=1}^n$ and unlabeled data: $\{\x_i\}_{i=n+1}^{N}$
  \begin{algorithmic}[1]
    \STATE {\bf\emph{Generate features.}}
	Draw $\omega_1,\ldots \omega_{2K}$ i.i.d. from $P$ and featurize the input: 
		\begin{align*}
		    \quad \z^{(1)}_i\leftarrow & \sq{\phi(\x_i;\boldomega_1),\ldots, \phi(\x_i;\boldomega_{M})}, \\
		    \quad \z^{(2)}_i\leftarrow & \sq{\phi(\x_i;\boldomega_{M+1}),\ldots, \phi(\x_i;\boldomega_{2M})}.
		\end{align*}
		\STATE {\bf \emph{Unlabeled data.}} Compute CCA bases $\Bt^{(1)}$, $\Bt^{(2)}$ and canonical correlations $\lambda_1,\ldots,\lambda_M$ for the two views and set
	    $\bar{\z}_i \leftarrow \Bt^{(1)}\z^{(1)}_i.$
 \STATE {\bf \emph{Labeled data.}} Solve
 \begin{equation}
\widehat{\wcca} =  \min_{\wcca} \frac{1}{n} \sum_{i=1}^n \loss \br{\wcca \tr \bar{\z}_i, y_i} + \Vert \wcca \Vert^2_{CCA} + \gamma \Vert \wcca \Vert_2^2 ~.
 \end{equation}    
	\end{algorithmic}
	\algorithmicensure\; $\widehat{\wcca}$
\end{algorithm}

It can be shown that, with sufficiently many features, views constructed via random Fourier features contain good approximations to a large class of functions with high probability, see main theorem of \cite{rahimi:08}. We do not provide details, since \cksrff is consistently outperformed by \cksny in practice.

%%%%%%%%%%%%%%%%%%%%%%%%%%%%%%%%%%%%%%%%%%%%%%%%%%%%%%%%%%%%
\FloatBarrier
\newpage
\addtocounter{si-sec}{1}
\section{Complete \cksrff results}
%%%%%%%%%%%%%%%%%%%%%%%%%%%%%%%%%%%%%%%%%%%%%%%%%%%%%%%%%%%%
For completeness we report on the performance of the \cksrff algorithm. We use the same experimental setup as in Section \ref{sec:results}. We compare the performance of \cksrff against a linear machine learned using $M$ and $2M$ random Fourier features respectively.

\begin{table}[htp]
\begin{center}
\caption{Average performance of \cksrff against \texttt{RFF}$_{M/2M}$ on 18 datasets.\label{tab:averagerff}}
\setlength{\tabcolsep}{3.5pt}
\small
\begin{tabular}{l r r r r r}
\hline
\cksrff vs \texttt{RFF}$_{M/2M}$ & $n=100$ & $n=200$ & $n=300$ & $n=400$ & $n=500$ \\
\hline
Avg reduction in error  &  15\%  & 30\%  &  34\% &  31\% & 28\%  \\ 
Avg reduction in std err   &  -1\%  & 35\%  &  47\% &  43\% & 44\% \\
\hline
\end{tabular}
\end{center}
\end{table}

\begin{figure}
\begin{centering}
\vspace{-10pt}
\subfloat[\texttt{abalone}]{\includegraphics[width=0.3\columnwidth]{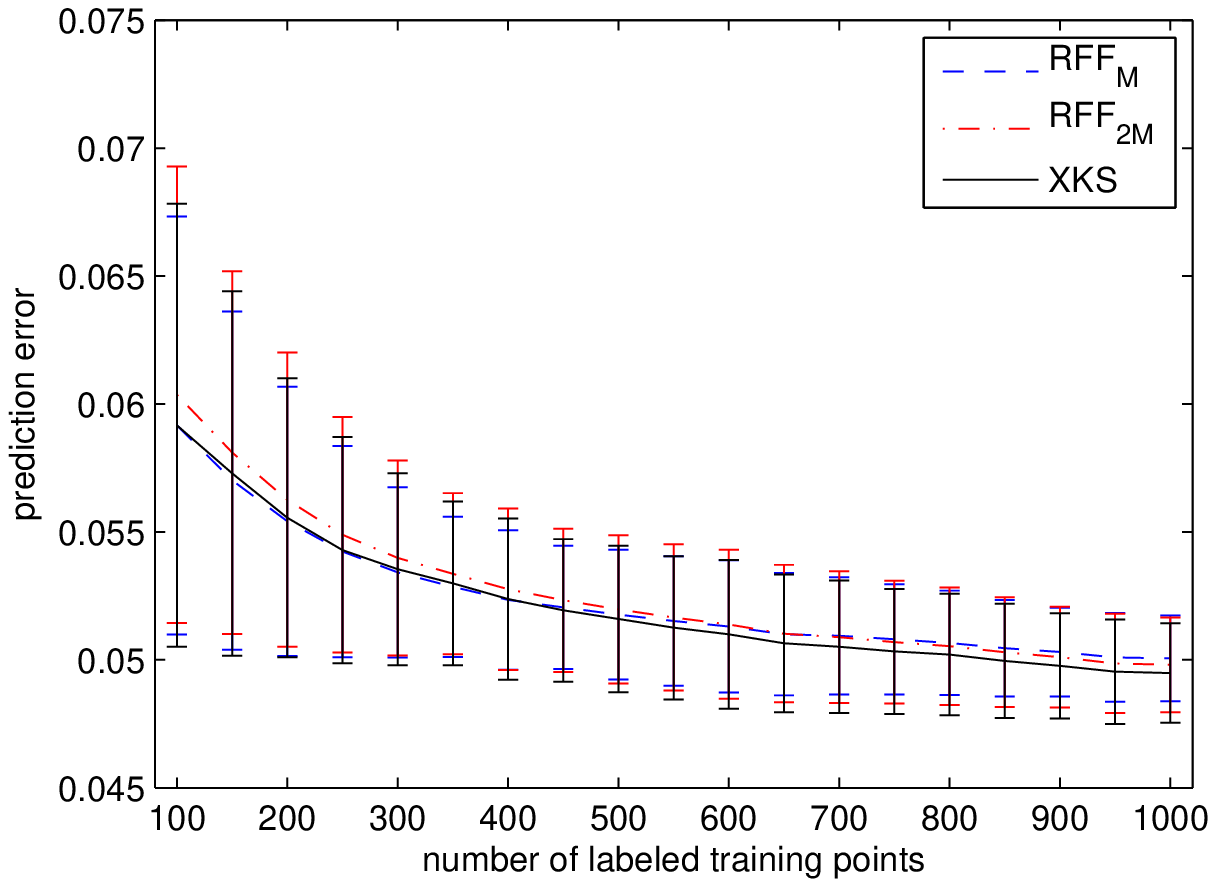} \label{fig:abalone}} 
\subfloat[\texttt{adult}]{\includegraphics[width=0.3\columnwidth]{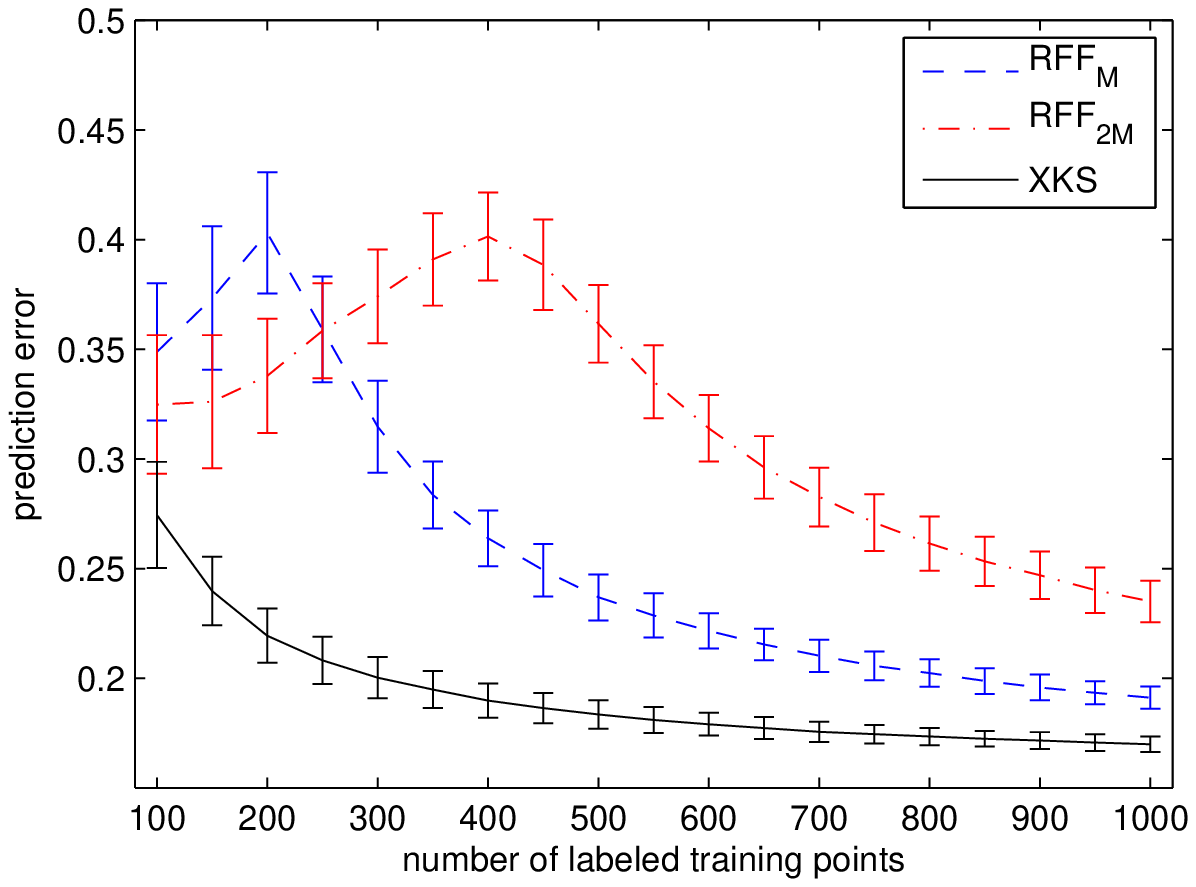} \label{fig:adult}} 
\subfloat[\texttt{ailerons}]{\includegraphics[width=0.3\columnwidth]{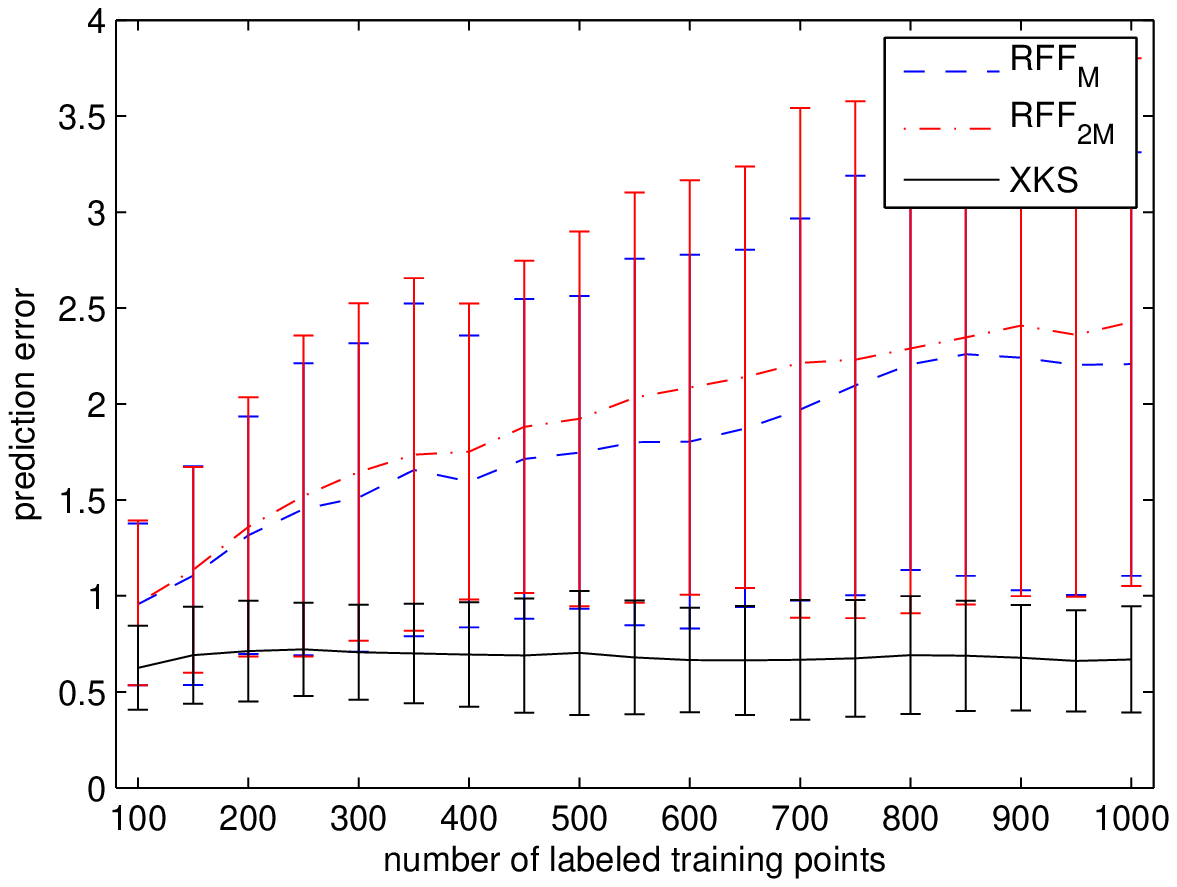} \label{fig:ailerons}} 
\vspace{-10pt}
\subfloat[\texttt{bank8}]{\includegraphics[width=0.3\columnwidth]{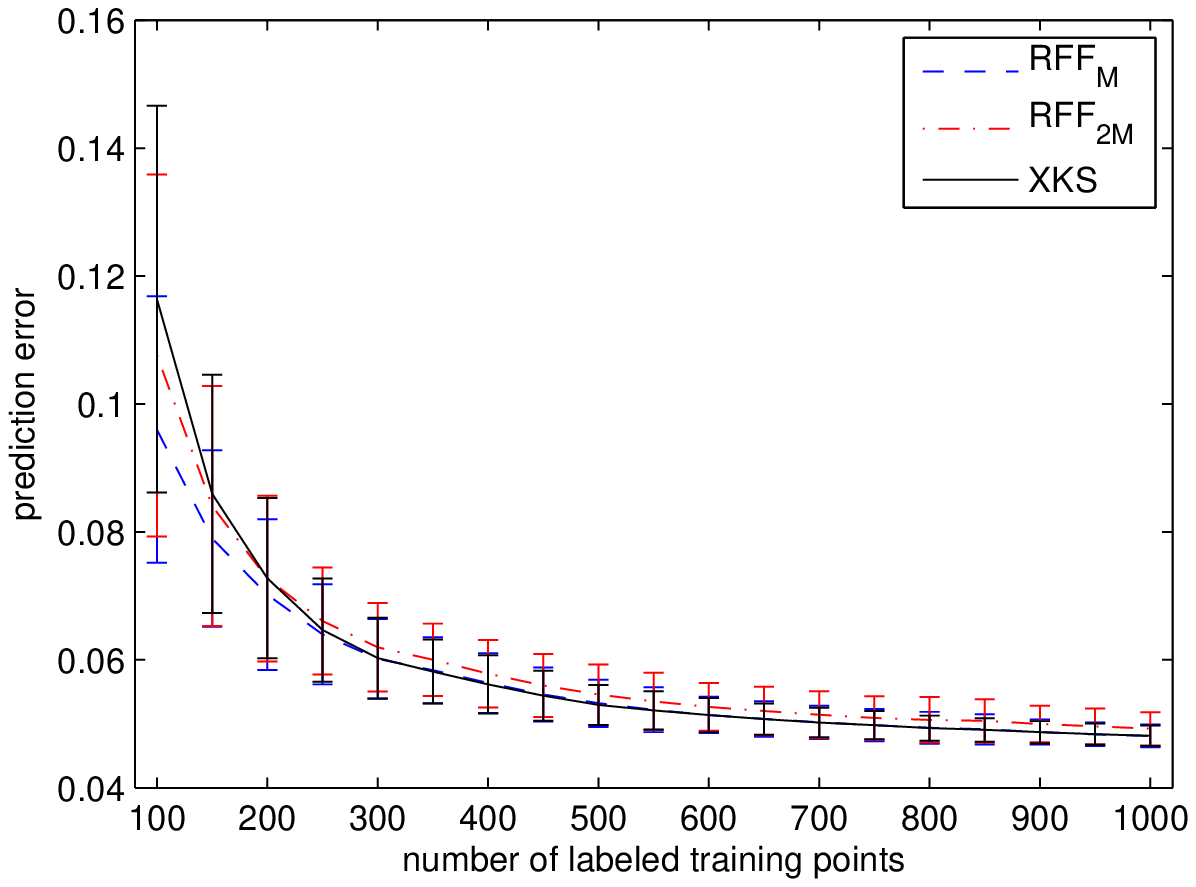} \label{fig:bank8}} 
\subfloat[\texttt{bank32}]{\includegraphics[width=0.3\columnwidth]{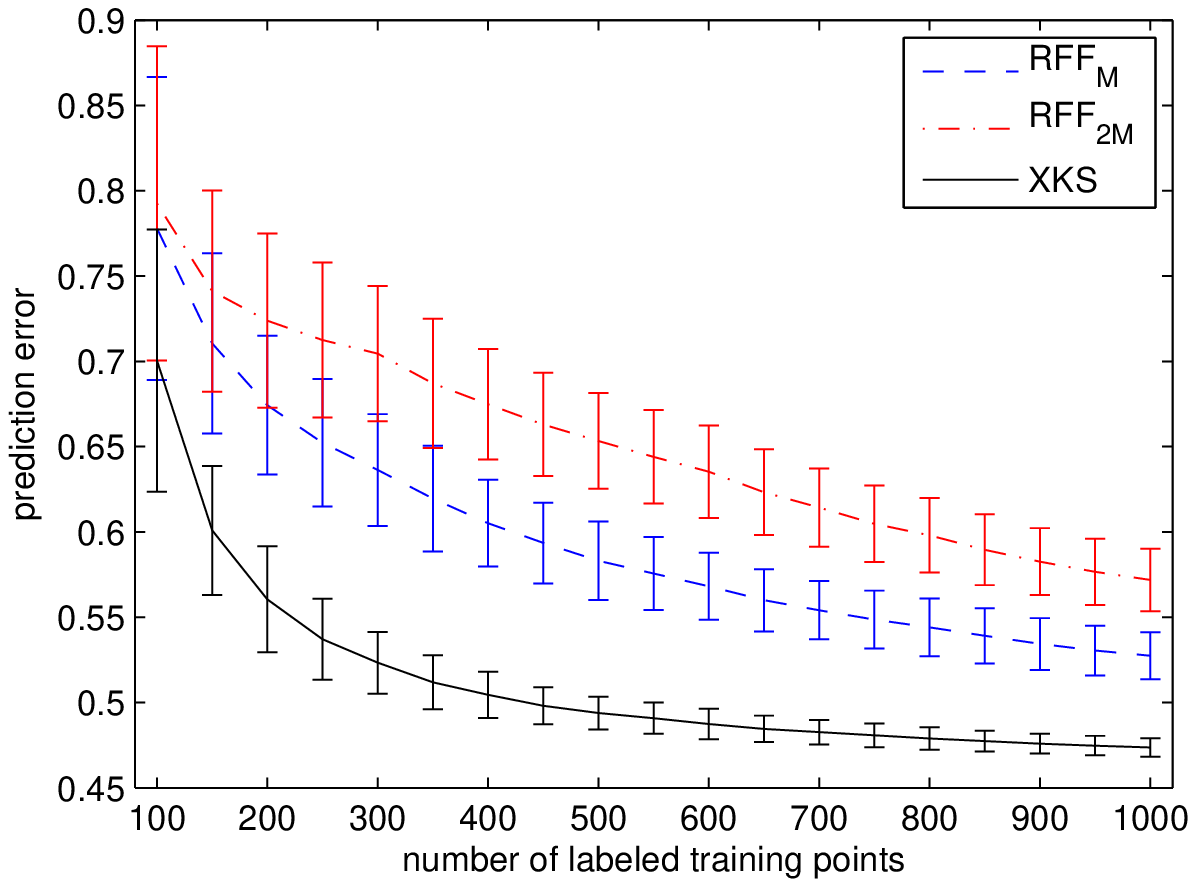} \label{fig:bank32}} 
\subfloat[\texttt{cal housing}]{\includegraphics[width=0.3\columnwidth]{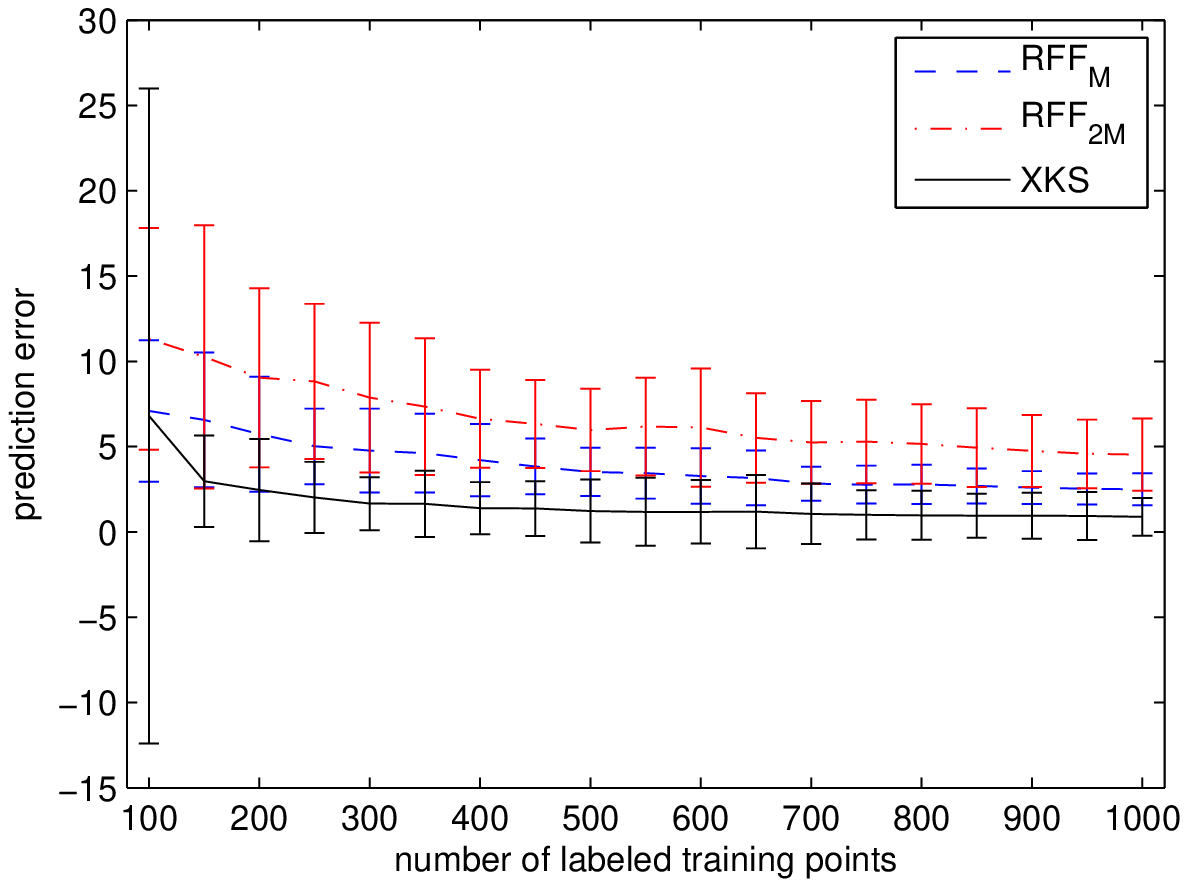} \label{fig:cal}}
\vspace{-10pt}
\subfloat[ \texttt{census}]{\includegraphics[width=0.3\textwidth]{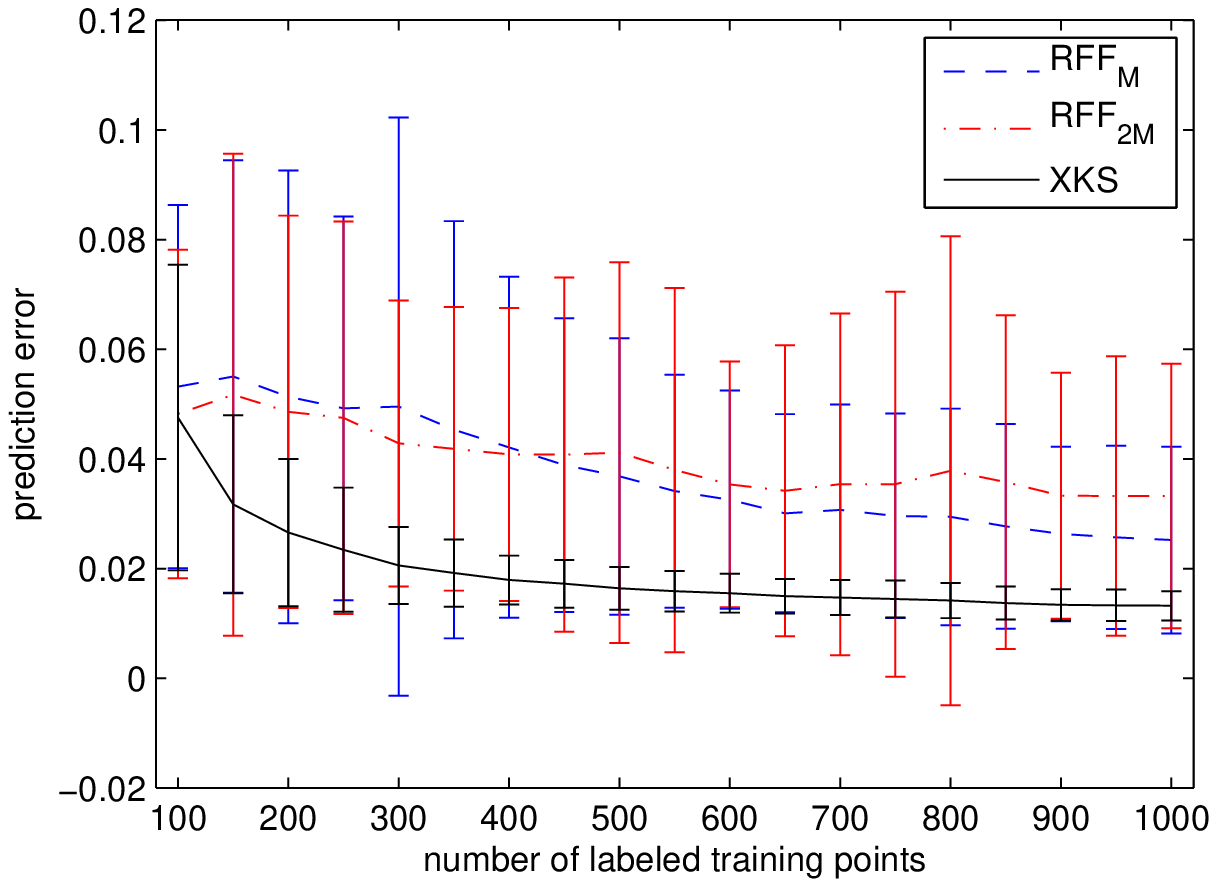} \label{fig:census}} 
\subfloat[ \texttt{CPU}]{\includegraphics[width=0.3\textwidth]{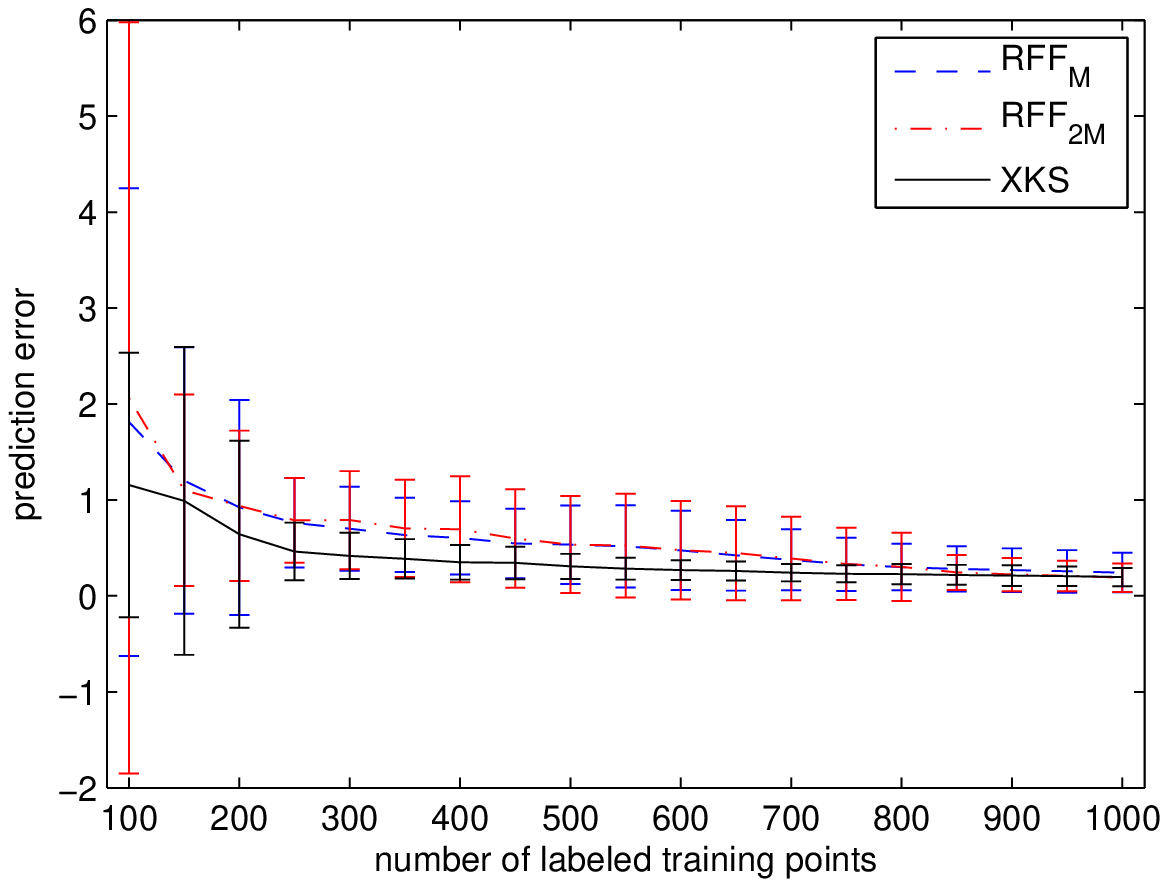} \label{fig:cpu}} 
\subfloat[ \texttt{CT}]{\includegraphics[width=0.3\textwidth]{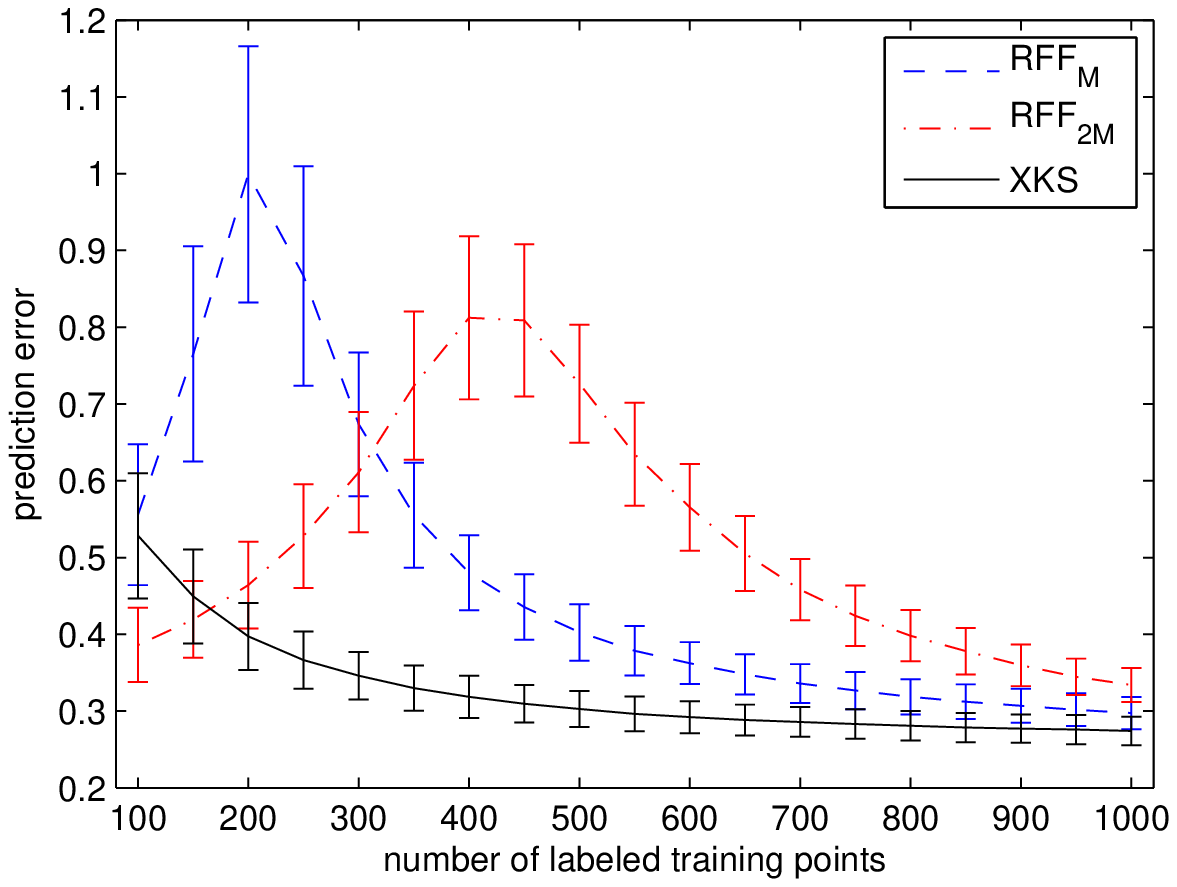} \label{fig:ct}}
\vspace{-10pt}
\subfloat[\texttt{elevators}]{\includegraphics[width=0.3\columnwidth]{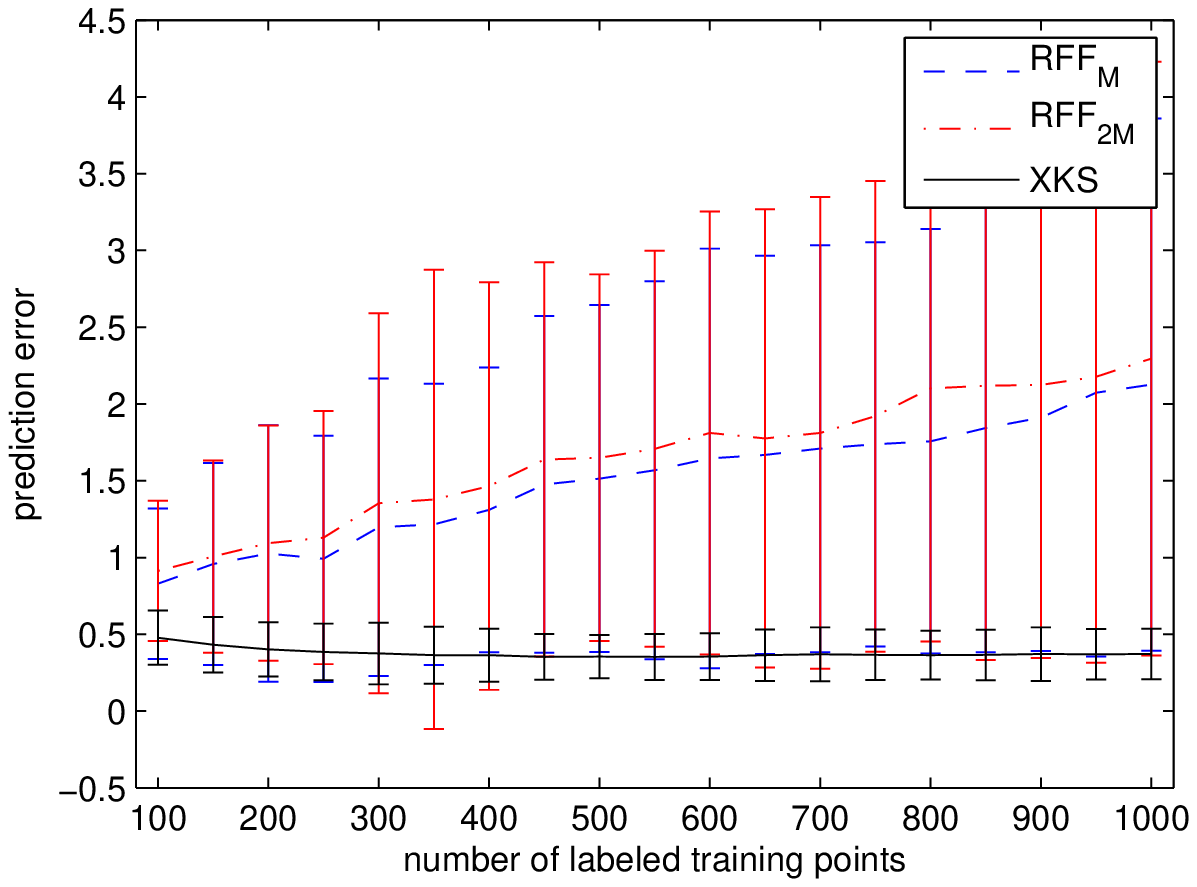} \label{fig:elevators}} 
\subfloat[\texttt{HIVa}]{\includegraphics[width=0.3\columnwidth]{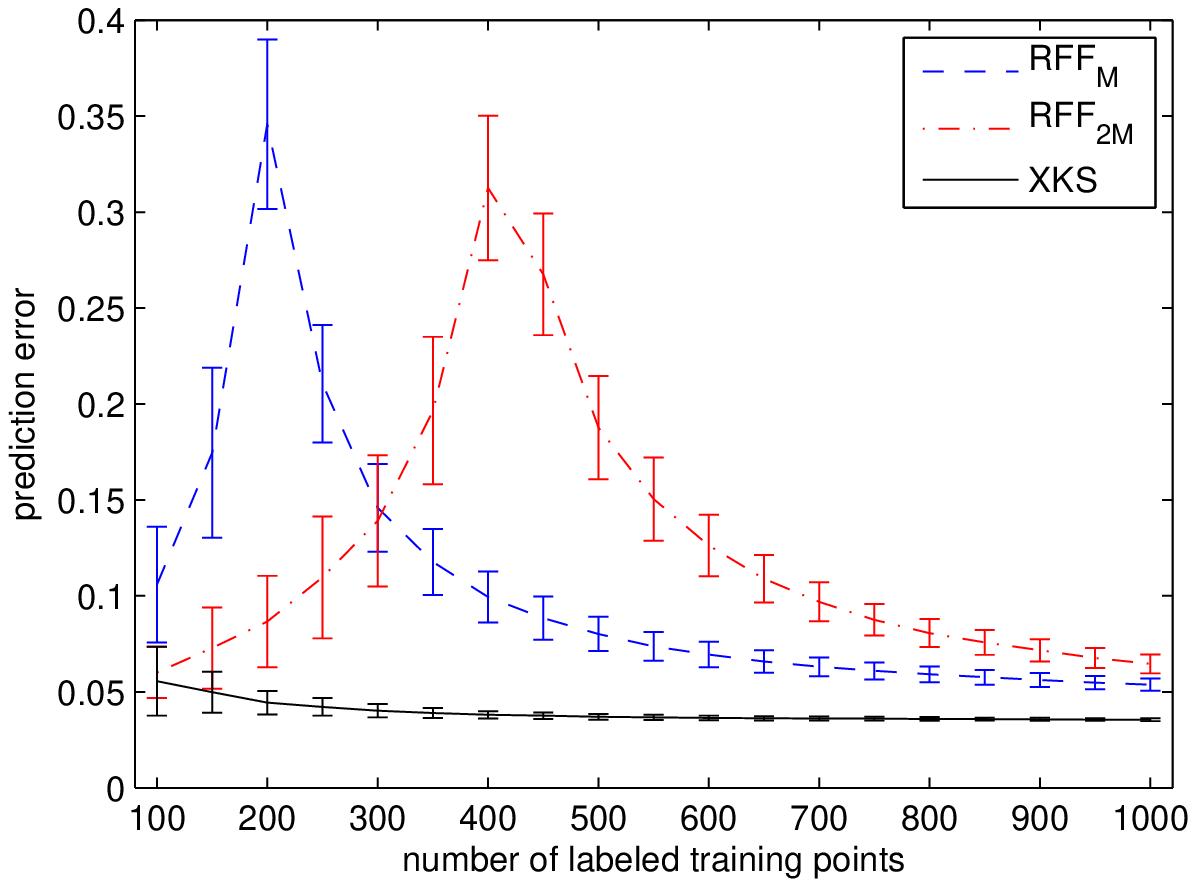} \label{fig:hiva}} 
\subfloat[\texttt{house}]{\includegraphics[width=0.3\columnwidth]{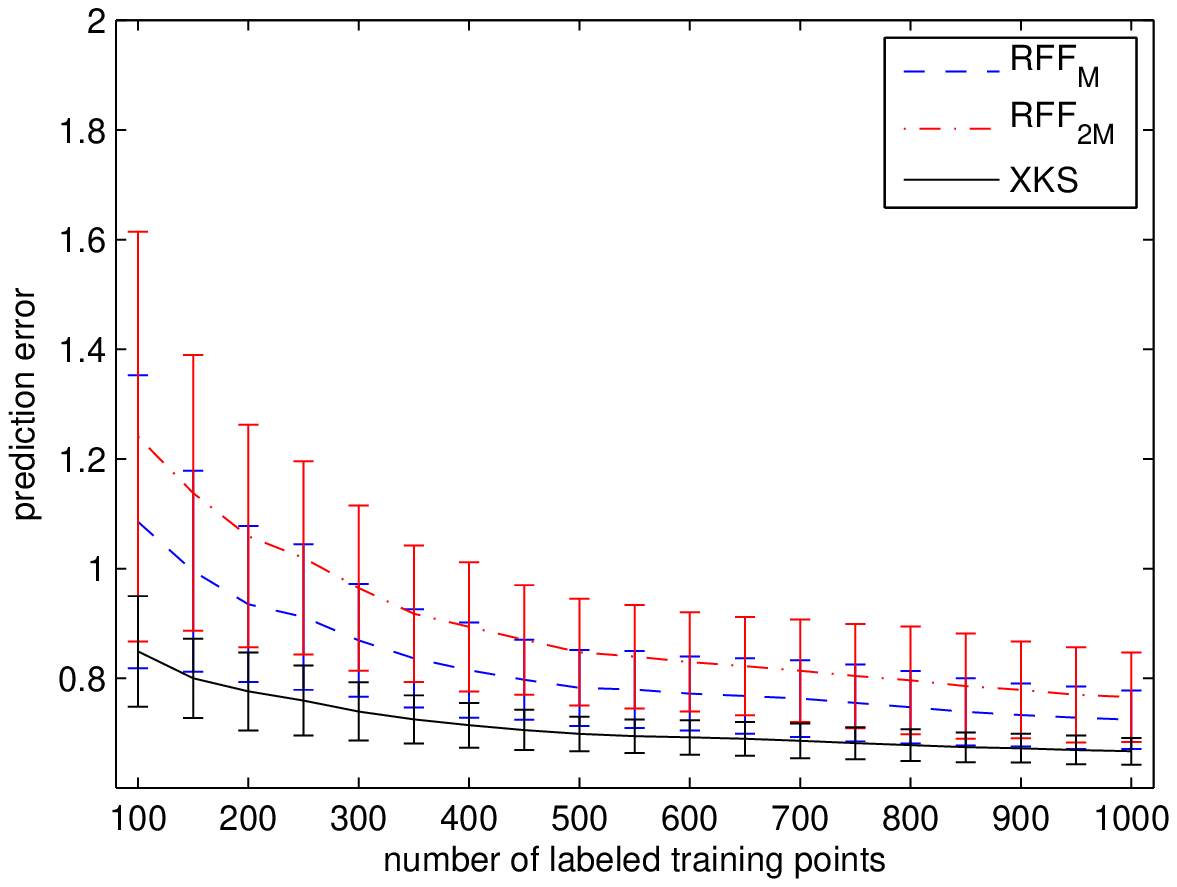} \label{fig:house}}
\vspace{-10pt}
\subfloat[\texttt{ibn Sina}]{\includegraphics[width=0.3\columnwidth]{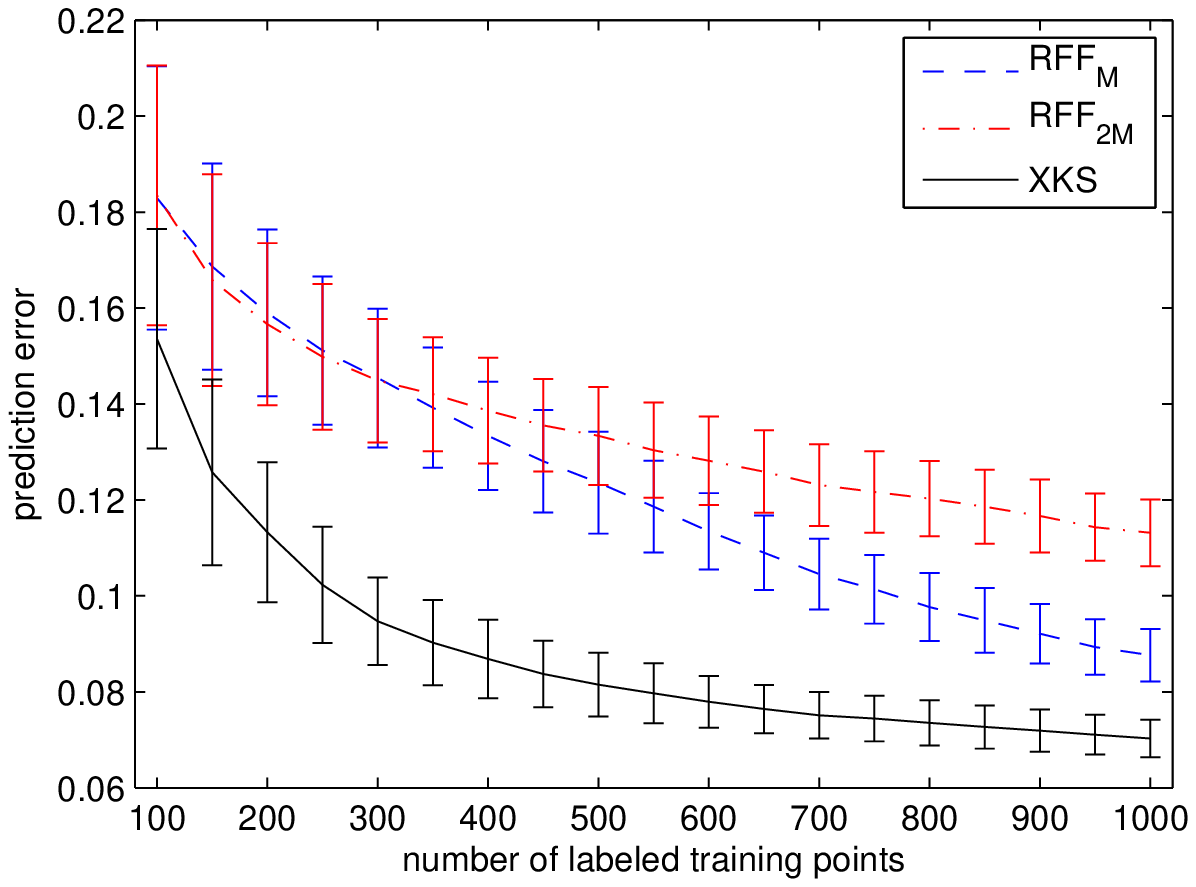} \label{fig:ibnsina}} 
\subfloat[\texttt{orange}]{\includegraphics[width=0.3\columnwidth]{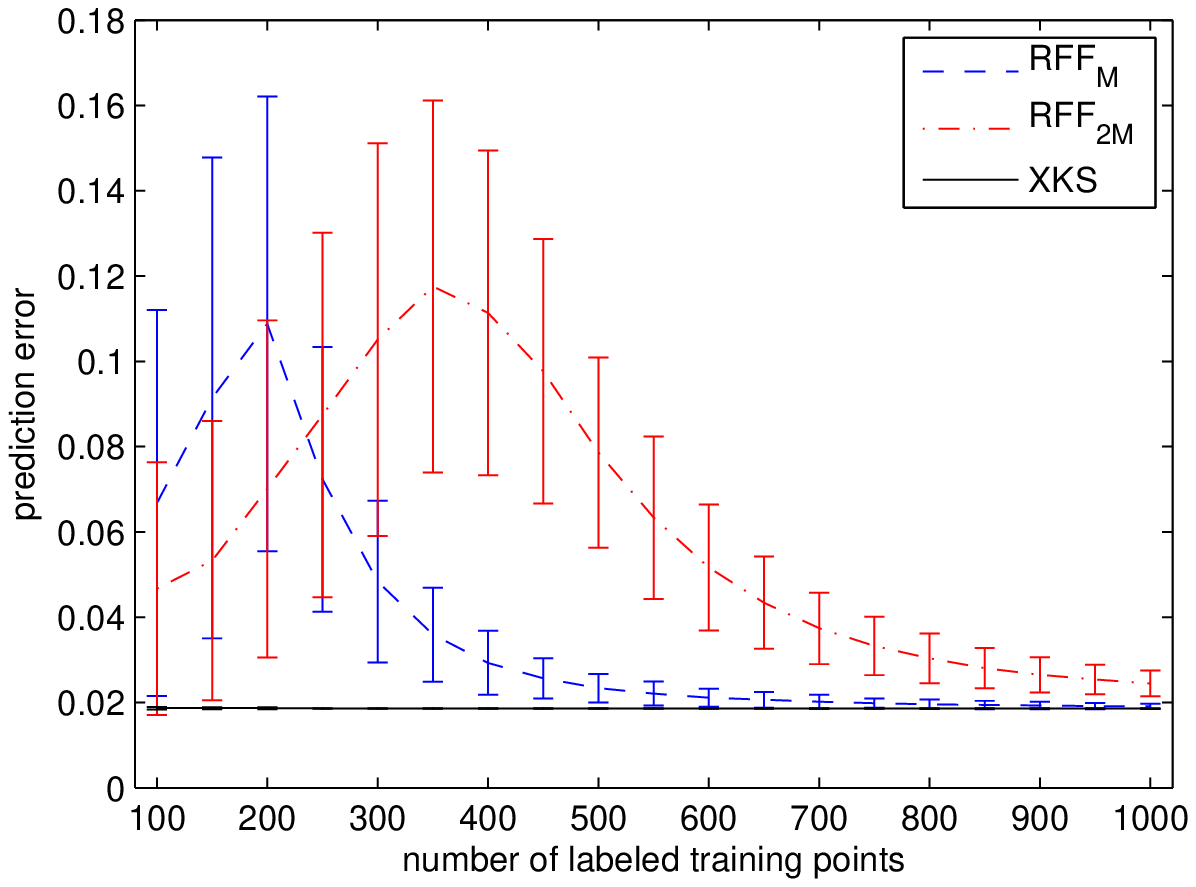} \label{fig:orange}} 
\subfloat[\texttt{sarcos 1}]{\includegraphics[width=0.3\columnwidth]{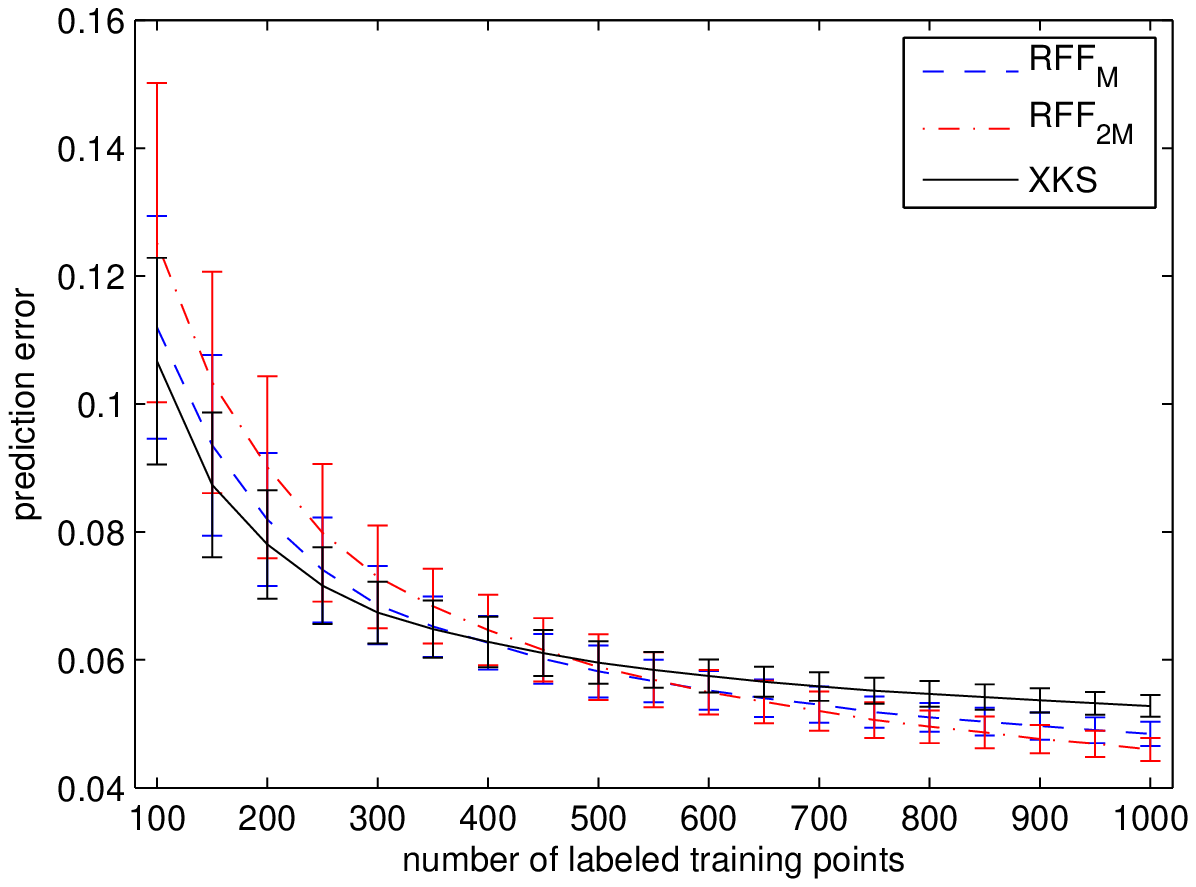} \label{fig:sarcos}}
\vspace{-10pt}
\subfloat[ \texttt{sarcos 5}]{\includegraphics[width=0.3\textwidth]{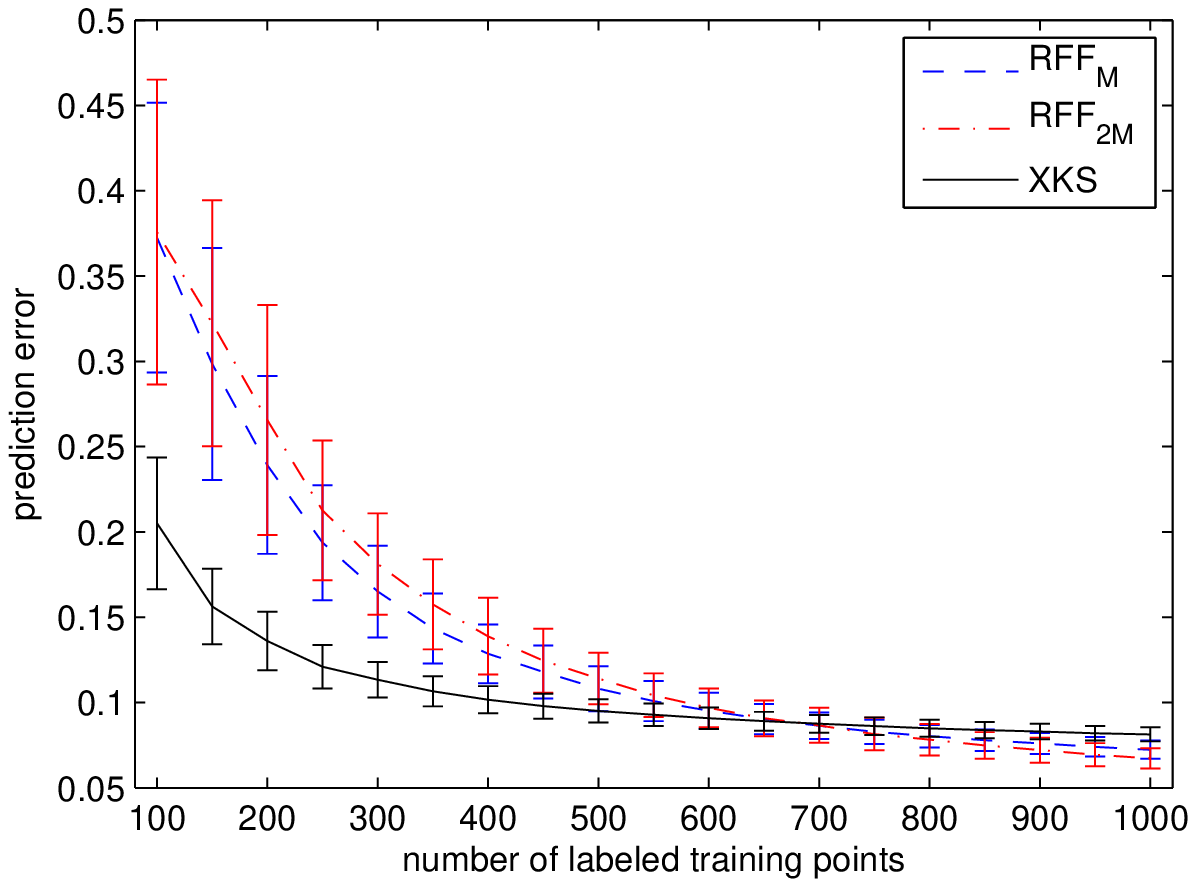} \label{fig:sarcos5}} 
\subfloat[ \texttt{sarcos 7}]{\includegraphics[width=0.3\textwidth]{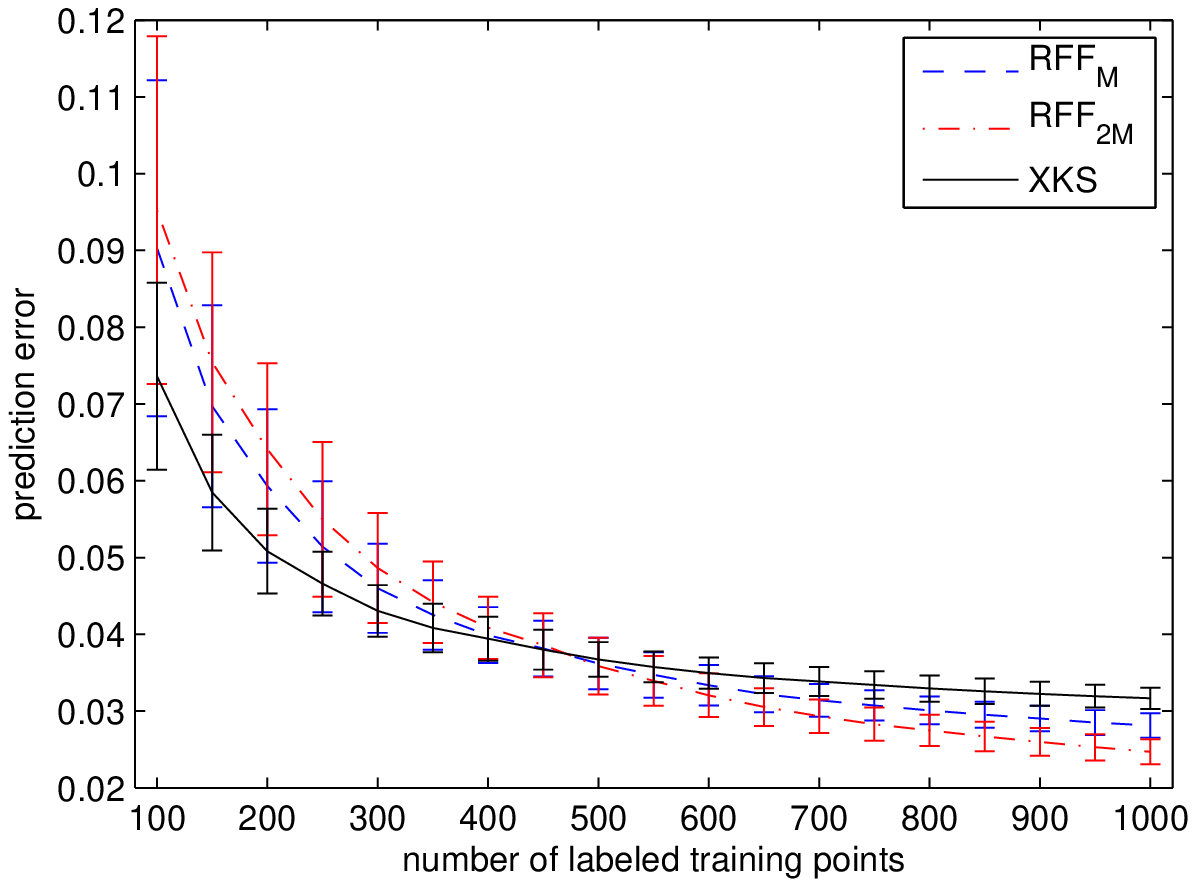} \label{fig:sarcos7}} 
\subfloat[ \texttt{sylva}]{\includegraphics[width=0.3\textwidth]{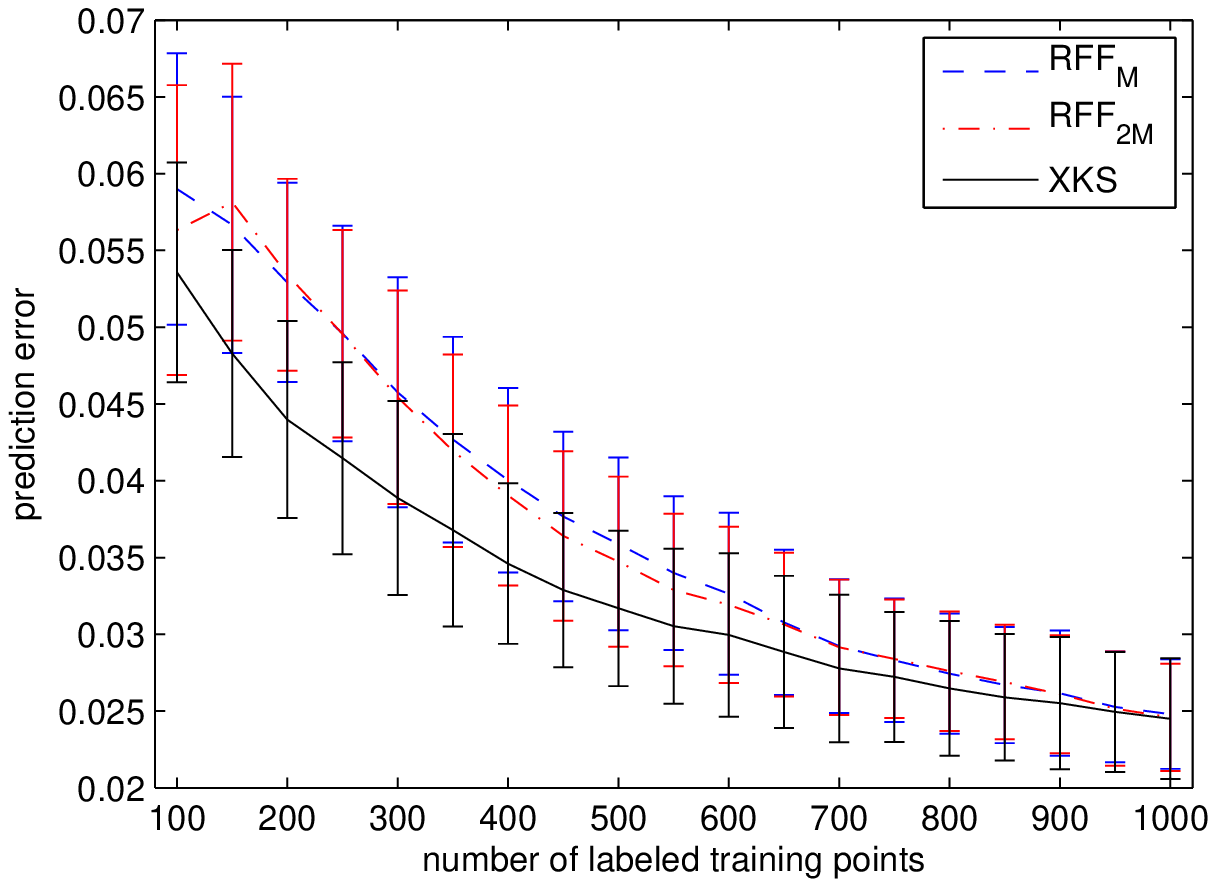} \label{fig:sylva}}  
\end{centering}
\vspace{-5pt}
\caption{Comparison of mean prediction error and standard deviation on all 18 datasets. \label{fig:res1rff}} 
\end{figure}

Table \ref{tab:averagerff} shows the performance improvement of \cksrff over \texttt{RFF}$_{M/2M}$, averaged across the 18 datasets. Table \ref{tab:results_rff1} compares the prediction error and standard deviation for each of the datasets individually. Figure \ref{fig:res1rff} shows the performance across the full range of values of $n$ for all datasets. The relative performance of \cksrff against \rffm and  \rffM follows the same trend seen in Section \ref{sec:results}, suggesting that CCA-based regression consistently improves on regression across single and joint views.

\begin{table}[htp]
\begin{center}
\caption{Number of datasets (out of 18) on which \cksny outperforms \cksrff.\label{tab:nywins}}
\setlength{\tabcolsep}{3.5pt}
\small
\begin{tabular}{l r r r r r}
\hline
& $n=100$ & $n=200$ & $n=300$ & $n=400$ & $n=500$ \\
\hline
&  16  & 16  &  15 &  16 & 16  \\ 
\hline
\end{tabular}
\end{center}
\end{table}

Finally, Table~\ref{tab:nywins} compares the performance of correlated Nystr\"om features against correlated kitchen sinks. \cksny typically outperforms \cksrff on 16 out of 18 datasets; with \cksrff only ever outperforming \cksny on \texttt{bank8}, \texttt{house} and \texttt{orange}. Since \cksny almost always outperforms \cksrff, we only discuss Nystr\"om features in the main text.

\begin{table}[htp]
\begin{center}
\caption{Performance of \cksrff (normalized MSE/classification error rate). Standard errors in parentheses. \label{tab:results_rff1}}
\setlength{\tabcolsep}{3.0pt}
\small
\begin{tabular}{l r r r | l r r r }
\hline
\multicolumn{1}{c}{set}&
\multicolumn{1}{c}{\rffm}&
\multicolumn{1}{c}{\rffM}&
\multicolumn{1}{c}{\cksrff}
 & set &
\multicolumn{1}{c}{\rffm}&
\multicolumn{1}{c}{\rffM}&
\multicolumn{1}{c}{\cksrff} \\ 
\hline 
\multicolumn{8}{l}{$n=100$}\\
\hline
1& ${\bf 0.059}$ $({\bf 0.008})$& $0.060$ $(0.009)$& $0.059$ $(0.009)$  & 10& $0.829$ $(0.490)$& $0.913$ $(0.457)$& ${\bf 0.478}$ $({\bf 0.176})$  \\ 
2& $0.349$ $(0.031)$& $0.325$ $(0.032)$& ${\bf 0.274}$ $({\bf 0.024})$  & 11& $0.106$ $(0.030)$& $0.060$ $(0.013)$& ${\bf 0.056}$ $({\bf 0.018})$  \\ 
3& $0.956$ $(0.421)$& $0.963$ $(0.428)$& ${\bf 0.626}$ $({\bf 0.220})$  & 12& $1.085$ $(0.267)$& $1.240$ $(0.374)$& ${\bf 0.849}$ $({\bf 0.101})$  \\ 
4& $0.778$ $(0.089)$& $0.793$ $(0.092)$& ${\bf 0.700}$ $({\bf 0.077})$  & 13& $0.183$ $(0.027)$& $0.183$ $(0.027)$& ${\bf 0.154}$ $({\bf 0.023})$  \\ 
5& ${\bf 0.096}$ $({\bf 0.021})$& $0.108$ $(0.028)$& $0.116$ $(0.030)$  & 14& $0.067$ $(0.045)$& $0.047$ $(0.030)$& ${\bf 0.019}$ $({\bf 0.000})$  \\ 
6& $7.091$ $(4.146)$& $11.320$ $(6.500)$& ${\bf 6.801}$ $({\bf 19.194})$  & 15& $0.112$ $(0.017)$& $0.125$ $(0.025)$& ${\bf 0.107}$ $({\bf 0.016})$  \\ 
7& $0.053$ $(0.033)$& $0.048$ $(0.030)$& ${\bf 0.048}$ $({\bf 0.028})$  & 16& $0.373$ $(0.079)$& $0.376$ $(0.089)$& ${\bf 0.205}$ $({\bf 0.039})$  \\ 
8& $1.813$ $(2.438)$& $2.062$ $(3.915)$& ${\bf 1.155}$ $({\bf 1.379})$  & 17& $0.090$ $(0.022)$& $0.095$ $(0.023)$& ${\bf 0.074}$ $({\bf 0.012})$  \\ 
9& $0.556$ $(0.092)$& ${\bf 0.386}$ $({\bf 0.048})$& $0.528$ $(0.082)$  & 18& $0.059$ $(0.009)$& $0.056$ $(0.009)$& ${\bf 0.054}$ $({\bf 0.007})$  \\ 
\hline
\multicolumn{8}{l}{$n=200$}\\
\hline
1& ${\bf 0.055}$ $({\bf 0.005})$& $0.056$ $(0.006)$& $0.056$ $(0.005)$  & 10& $1.026$ $(0.837)$& $1.094$ $(0.766)$& ${\bf 0.402}$ $({\bf 0.177})$  \\ 
2& $0.403$ $(0.028)$& $0.338$ $(0.026)$& ${\bf 0.219}$ $({\bf 0.012})$  & 11& $0.346$ $(0.044)$& $0.087$ $(0.024)$& ${\bf 0.044}$ $({\bf 0.006})$  \\ 
3& $1.316$ $(0.619)$& $1.359$ $(0.675)$& ${\bf 0.713}$ $({\bf 0.262})$  & 12& $0.935$ $(0.142)$& $1.059$ $(0.203)$& ${\bf 0.776}$ $({\bf 0.071})$  \\ 
4& $0.674$ $(0.041)$& $0.724$ $(0.051)$& ${\bf 0.561}$ $({\bf 0.031})$  & 13& $0.159$ $(0.017)$& $0.157$ $(0.017)$& ${\bf 0.113}$ $({\bf 0.015})$  \\ 
5& ${\bf 0.070}$ $({\bf 0.012})$& $0.073$ $(0.013)$& $0.073$ $(0.013)$  & 14& $0.109$ $(0.053)$& $0.070$ $(0.040)$& ${\bf 0.019}$ $({\bf 0.000})$  \\ 
6& $5.731$ $(3.367)$& $9.037$ $(5.248)$& ${\bf 2.454}$ $({\bf 2.998})$  & 15& $0.082$ $(0.010)$& $0.090$ $(0.014)$& ${\bf 0.078}$ $({\bf 0.008})$  \\ 
7& $0.051$ $(0.041)$& $0.049$ $(0.036)$& ${\bf 0.027}$ $({\bf 0.013})$  & 16& $0.239$ $(0.052)$& $0.266$ $(0.067)$& ${\bf 0.136}$ $({\bf 0.017})$  \\ 
8& $0.922$ $(1.119)$& $0.938$ $(0.783)$& ${\bf 0.643}$ $({\bf 0.974})$  & 17& $0.059$ $(0.010)$& $0.064$ $(0.011)$& ${\bf 0.051}$ $({\bf 0.006})$  \\ 
9& $0.999$ $(0.167)$& $0.464$ $(0.057)$& ${\bf 0.397}$ $({\bf 0.043})$  & 18& $0.053$ $(0.006)$& $0.053$ $(0.006)$& ${\bf 0.044}$ $({\bf 0.006})$  \\ 
\hline
\multicolumn{8}{l}{$n=300$}\\
\hline
1& ${\bf 0.053}$ $({\bf 0.003})$& $0.054$ $(0.004)$& $0.054$ $(0.004)$  & 10& $1.197$ $(0.969)$& $1.354$ $(1.238)$& ${\bf 0.375}$ $({\bf 0.201})$  \\ 
2& $0.315$ $(0.021)$& $0.374$ $(0.021)$& ${\bf 0.200}$ $({\bf 0.009})$  & 11& $0.146$ $(0.023)$& $0.139$ $(0.034)$& ${\bf 0.040}$ $({\bf 0.003})$  \\ 
3& $1.513$ $(0.804)$& $1.646$ $(0.878)$& ${\bf 0.706}$ $({\bf 0.248})$  & 12& $0.869$ $(0.103)$& $0.964$ $(0.151)$& ${\bf 0.739}$ $({\bf 0.053})$  \\ 
4& $0.636$ $(0.033)$& $0.705$ $(0.040)$& ${\bf 0.523}$ $({\bf 0.018})$  & 13& $0.145$ $(0.014)$& $0.145$ $(0.013)$& ${\bf 0.095}$ $({\bf 0.009})$  \\ 
5& ${\bf 0.060}$ $({\bf 0.006})$& $0.062$ $(0.007)$& $0.060$ $(0.006)$  & 14& $0.048$ $(0.019)$& $0.105$ $(0.046)$& ${\bf 0.019}$ $({\bf 0.000})$  \\ 
6& $4.769$ $(2.468)$& $7.871$ $(4.393)$& ${\bf 1.660}$ $({\bf 1.549})$  & 15& $0.069$ $(0.006)$& $0.073$ $(0.008)$& ${\bf 0.067}$ $({\bf 0.005})$  \\ 
7& $0.050$ $(0.053)$& $0.043$ $(0.026)$& ${\bf 0.021}$ $({\bf 0.007})$  & 16& $0.165$ $(0.027)$& $0.181$ $(0.030)$& ${\bf 0.113}$ $({\bf 0.010})$  \\ 
8& $0.699$ $(0.437)$& $0.789$ $(0.511)$& ${\bf 0.416}$ $({\bf 0.241})$  & 17& $0.046$ $(0.006)$& $0.049$ $(0.007)$& ${\bf 0.043}$ $({\bf 0.003})$  \\ 
9& $0.673$ $(0.094)$& $0.611$ $(0.078)$& ${\bf 0.346}$ $({\bf 0.031})$  & 18& $0.046$ $(0.007)$& $0.045$ $(0.007)$& ${\bf 0.039}$ $({\bf 0.006})$  \\ 
\hline
\multicolumn{8}{l}{$n=400$}\\
\hline
1& ${\bf 0.052}$ $({\bf 0.003})$& $0.053$ $(0.003)$& $0.052$ $(0.003)$  & 10& $1.311$ $(0.927)$& $1.466$ $(1.328)$& ${\bf 0.364}$ $({\bf 0.172})$  \\ 
2& $0.264$ $(0.013)$& $0.401$ $(0.020)$& ${\bf 0.190}$ $({\bf 0.008})$  & 11& $0.099$ $(0.013)$& $0.313$ $(0.038)$& ${\bf 0.038}$ $({\bf 0.002})$  \\ 
3& $1.596$ $(0.760)$& $1.752$ $(0.771)$& ${\bf 0.695}$ $({\bf 0.273})$  & 12& $0.815$ $(0.087)$& $0.894$ $(0.118)$& ${\bf 0.714}$ $({\bf 0.041})$  \\ 
4& $0.605$ $(0.025)$& $0.675$ $(0.032)$& ${\bf 0.504}$ $({\bf 0.014})$  & 13& $0.133$ $(0.011)$& $0.139$ $(0.011)$& ${\bf 0.087}$ $({\bf 0.008})$  \\ 
5& $0.056$ $(0.005)$& $0.058$ $(0.005)$& ${\bf 0.056}$ $({\bf 0.005})$  & 14& $0.029$ $(0.007)$& $0.111$ $(0.038)$& ${\bf 0.019}$ $({\bf 0.000})$  \\ 
6& $4.214$ $(2.123)$& $6.632$ $(2.862)$& ${\bf 1.394}$ $({\bf 1.533})$  & 15& ${\bf 0.063}$ $({\bf 0.004})$& $0.065$ $(0.006)$& $0.063$ $(0.004)$  \\ 
7& $0.042$ $(0.031)$& $0.041$ $(0.027)$& ${\bf 0.018}$ $({\bf 0.004})$  & 16& $0.129$ $(0.017)$& $0.139$ $(0.022)$& ${\bf 0.102}$ $({\bf 0.008})$  \\ 
8& $0.605$ $(0.382)$& $0.695$ $(0.553)$& ${\bf 0.350}$ $({\bf 0.181})$  & 17& $0.040$ $(0.004)$& $0.041$ $(0.004)$& ${\bf 0.039}$ $({\bf 0.003})$  \\ 
9& $0.480$ $(0.049)$& $0.812$ $(0.106)$& ${\bf 0.318}$ $({\bf 0.027})$  & 18& $0.040$ $(0.006)$& $0.039$ $(0.006)$& ${\bf 0.035}$ $({\bf 0.005})$  \\ 
\hline
\multicolumn{8}{l}{$n=500$}\\
\hline
1& $0.052$ $(0.003)$& $0.052$ $(0.003)$& ${\bf 0.052}$ $({\bf 0.003})$  & 10& $1.514$ $(1.130)$& $1.650$ $(1.195)$& ${\bf 0.355}$ $({\bf 0.142})$  \\ 
2& $0.237$ $(0.010)$& $0.362$ $(0.018)$& ${\bf 0.183}$ $({\bf 0.006})$  & 11& $0.080$ $(0.009)$& $0.188$ $(0.027)$& ${\bf 0.037}$ $({\bf 0.002})$  \\ 
3& $1.747$ $(0.815)$& $1.923$ $(0.976)$& ${\bf 0.703}$ $({\bf 0.323})$  & 12& $0.782$ $(0.069)$& $0.847$ $(0.097)$& ${\bf 0.698}$ $({\bf 0.031})$  \\ 
4& $0.583$ $(0.023)$& $0.653$ $(0.028)$& ${\bf 0.494}$ $({\bf 0.010})$  & 13& $0.124$ $(0.011)$& $0.133$ $(0.010)$& ${\bf 0.082}$ $({\bf 0.007})$  \\ 
5& $0.053$ $(0.004)$& $0.055$ $(0.005)$& ${\bf 0.053}$ $({\bf 0.003})$  & 14& $0.023$ $(0.003)$& $0.079$ $(0.022)$& ${\bf 0.019}$ $({\bf 0.000})$  \\ 
6& $3.515$ $(1.416)$& $5.977$ $(2.419)$& ${\bf 1.231}$ $({\bf 1.848})$  & 15& ${\bf 0.058}$ $({\bf 0.004})$& $0.059$ $(0.005)$& $0.060$ $(0.003)$  \\ 
7& $0.037$ $(0.025)$& $0.041$ $(0.035)$& ${\bf 0.016}$ $({\bf 0.004})$  & 16& $0.108$ $(0.013)$& $0.114$ $(0.015)$& ${\bf 0.095}$ $({\bf 0.007})$  \\ 
8& $0.533$ $(0.408)$& $0.536$ $(0.505)$& ${\bf 0.307}$ $({\bf 0.132})$  & 17& $0.036$ $(0.003)$& ${\bf 0.036}$ $({\bf 0.004})$& $0.037$ $(0.002)$  \\ 
9& $0.403$ $(0.037)$& $0.726$ $(0.077)$& ${\bf 0.303}$ $({\bf 0.023})$  & 18& $0.036$ $(0.006)$& $0.035$ $(0.006)$& ${\bf 0.032}$ $({\bf 0.005})$  \\ 
\hline
\end{tabular}
\end{center}
\end{table}

%%%%%%% prediction error plots set 3
%\begin{figure}
%\begin{centering}
%\subfloat[\texttt{ibn Sina}]{\includegraphics[width=0.33\columnwidth]{rff/eps/19_ibn_sina} \label{fig:ibnsina}} 
%\subfloat[\texttt{orange}]{\includegraphics[width=0.33\columnwidth]{rff/eps/19_orange} \label{fig:orange}} 
%\subfloat[\texttt{sarcos 1}]{\includegraphics[width=0.33\columnwidth]{rff/eps/19_sarcos} \label{fig:sarcos}}
%\\ 
%\subfloat[ \texttt{sarcos 5}]{\includegraphics[width=0.33\textwidth]{rff/eps/19_sarcos5} \label{fig:sarcos5}} 
%\subfloat[ \texttt{sarcos 7}]{\includegraphics[width=0.33\textwidth]{rff/eps/19_sarcos7} \label{fig:sarcos7}} 
%\subfloat[ \texttt{sylva}]{\includegraphics[width=0.33\textwidth]{rff/eps/19_sylva} \label{fig:sylva}} 
%\caption{Mean prediction error and standard deviation. \label{fig:res3rff}}
%\end{centering}
%\end{figure}

%\input{supp_plots} 
\end{document}